\documentclass{article}

    \usepackage[preprint]{neurips_2025}

\usepackage{graphicx}
\usepackage[utf8]{inputenc} 
\usepackage[T1]{fontenc}    
\usepackage{url}            
\usepackage{booktabs}       
\usepackage{amsfonts}       
\usepackage{nicefrac}       
\usepackage{microtype}      
\usepackage{xcolor}         
\usepackage{wrapfig,lipsum,booktabs}
\usepackage{multirow}
\usepackage{subcaption}
\usepackage{csquotes}
\usepackage{amsmath}
\usepackage{tcolorbox}

\usepackage{listings}

\usepackage{algpseudocode}
\usepackage{algorithm}

\definecolor{citationblue}{rgb}{0.21,0.49,0.74}
\definecolor{limegreen}{rgb}{0.2, 0.8, 0.2}
\definecolor{ered}{rgb}{0.72, 0.16, 0.2}
\usepackage[pagebackref,breaklinks,colorlinks,allcolors=citationblue]{hyperref}

\newcommand{\wrapassumpbox}[5]{%
  \begin{wrapfigure}[#1]{#2}{#3}
    \vspace{-4ex}
    \begin{tcolorbox}[colback=white!98!black,
                      colframe=white!30!black,
                      boxsep=1.1pt,
                      top=6.75pt,
                      left=2pt,
                      right=2pt]
      \textbf{#4}\\[-0.575em]
      \noindent\rule{\linewidth}{0.4pt}\\[0.25em]
      #5
    \end{tcolorbox}
  \end{wrapfigure}
}

\newcommand{\q}[1]{``#1''} 

\newcommand{\improve}[1]{\tiny\textcolor{limegreen}{\boldmath $(+ #1)$}}

\usepackage[capitalize,noabbrev]{cleveref}

\AtEndPreamble{
    \usepackage[capitalize]{cleveref}
    \crefname{section}{Sec.}{Secs.}
    \Crefname{section}{Section}{Sections}
    \Crefname{table}{Table}{Tables}
    \crefname{table}{Tab.}{Tabs.}
    \Crefname{equation}{Eq.}{Eqs.}
    \Crefname{figure}{Fig.}{Figs.}
    \Crefname{tabular}{Tab.}{Tabs.}
    \Crefname{algorithm}{Alg.}{Algs.}
    \Crefname{proposition}{Prop.}{Props.}
    \Crefname{appendix}{App.}{Apps.}
    \creflabelformat{equation}{#2\textup{#1}#3}
}

\title{We Should Chart an Atlas of All the World's Models}

\author{Eliahu Horwitz \qquad Nitzan Kurer \qquad Jonathan Kahana \qquad Liel Amar \qquad Yedid Hoshen \\
School of Computer Science and Engineering\\
The Hebrew University of Jerusalem, Israel\\
\small\url{https://horwitz.ai/model-atlas}\\
}

\begin{document}

\maketitle

\begin{figure}[h]
\centering
    \includegraphics[width=\linewidth]{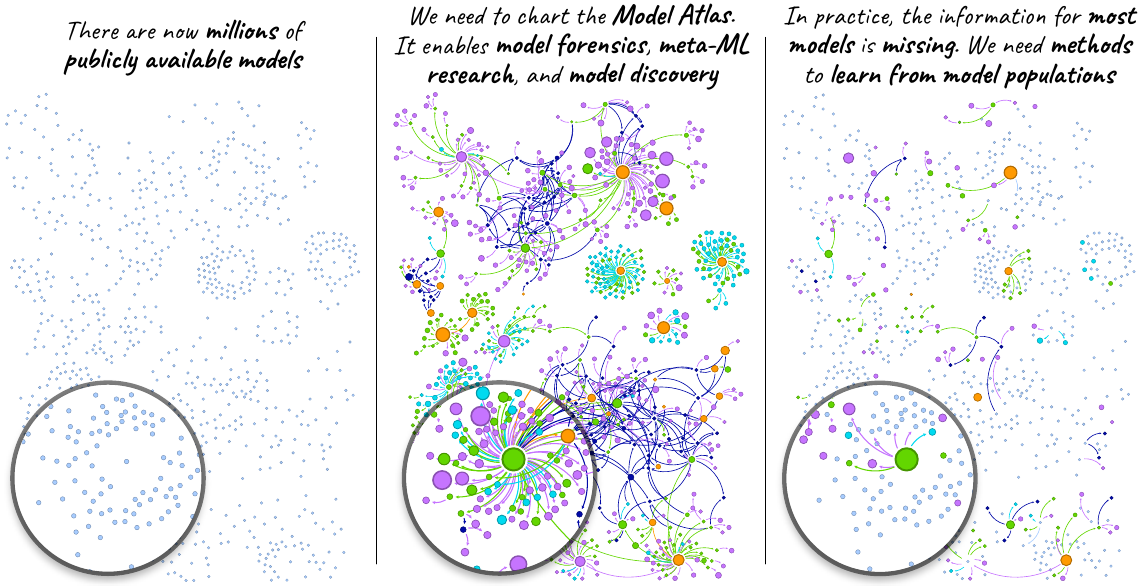} 
\caption{\textit{\textbf{Position overview:}} With millions of public models, it becomes important to move beyond individual models and study entire populations (left). The Model Atlas formalizes this shift by representing models as nodes in a graph, with directed edges denoting weight transformations (e.g., fine-tuning). Node size and color, as well as edge color, encode node and edge-level features; light blue indicates missing or unknown information. The atlas enables a range of applications, including model forensics, meta-ML research, and model discovery (center). In practice, most edges and features are unknown. This motivates ML methods that take models as input and infer their properties, thereby completing the missing atlas regions (right). \textbf{Zoom in to view edges, best viewed in color.}}
 \label{fig:teaser}
\end{figure}

\begin{abstract}
Public model repositories now contain millions of models, yet most models remain undocumented and effectively lost. In this position paper, we advocate for charting the world’s model population in a unified structure we call the \textit{Model Atlas}: a graph that captures models, their attributes, and the weight transformations that connect them. The Model Atlas enables applications in model forensics, meta-ML research, and model discovery, challenging tasks given today’s unstructured model repositories. However, because most models lack documentation, large atlas regions remain uncharted. Addressing this gap motivates new machine learning methods that treat models themselves as data, inferring properties such as functionality, performance, and lineage directly from their weights. We argue that a scalable path forward is to bypass the unique parameter symmetries that plague model weights. Charting all the world’s models will require a community effort, and we hope its broad utility will rally researchers toward this goal.
\end{abstract}

\section{Introduction}
\label{sec:intro}
Many scientific breakthroughs occurred when researchers transitioned from observing individual samples to studying entire populations. Consider Darwin's transition from cataloging specimens to understanding evolution through natural selection acting on populations, or modern medicine's move from anecdotal patient outcomes to large-scale clinical trials establishing treatment efficacy. Machine learning (ML) has mirrored this shift for data, replacing hand-crafted features derived from a few observations with representations learned from large sample collections. This paradigm has fueled remarkable progress in many domains like computer vision (CV), natural language processing (NLP), and audio. However, we argue that the ML community has yet to fully apply this population-level perspective to its outputs: the models themselves. While we optimize individual models, we have largely ignored the implicit knowledge embedded across the growing population of trained models. \textbf{In this position paper, we advocate for systematically studying entire model populations, and argue that this requires charting them in a unified structure, the \textit{\q{Model Atlas}}.} We present a schematic overview of this position in \cref{fig:teaser}.

\begin{wrapfigure}[12]{R}{0.5\textwidth}
\centering
\includegraphics[width=0.5\textwidth]{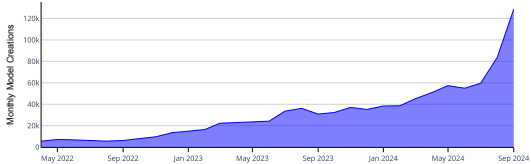}
\caption{\textit{\textbf{Growth in Hugging Face models:}} The number of public models is growing rapidly, but most remain undocumented and effectively lost. We advocate for charting them in a Model Atlas.}
\label{fig:hf_growth}
\end{wrapfigure}

This direction is particularly timely: the number of publicly available models is growing at an unprecedented rate. Platforms like Hugging Face (HF) now host over 1.5 million models, with over 100k added each month (see \cref{fig:hf_growth}). Yet, most models remain undocumented and effectively lost. We therefore need to chart them, mapping the machine learning landscape. Concretely, we define the Model Atlas as a graph where nodes represent models and directed edges capture weight transformations from one model to another (e.g., fine-tuning, quantization, merging). Additionally, nodes and edges are labeled with attributes such as task performance, training data, and optimization details. The atlas transforms model collections from a flat list of models to an interconnected ecosystem. This enables a range of applications, e.g., model forensics, meta-ML research, and model discovery.

\begin{figure}[t]
\centering
\includegraphics[width=\linewidth]{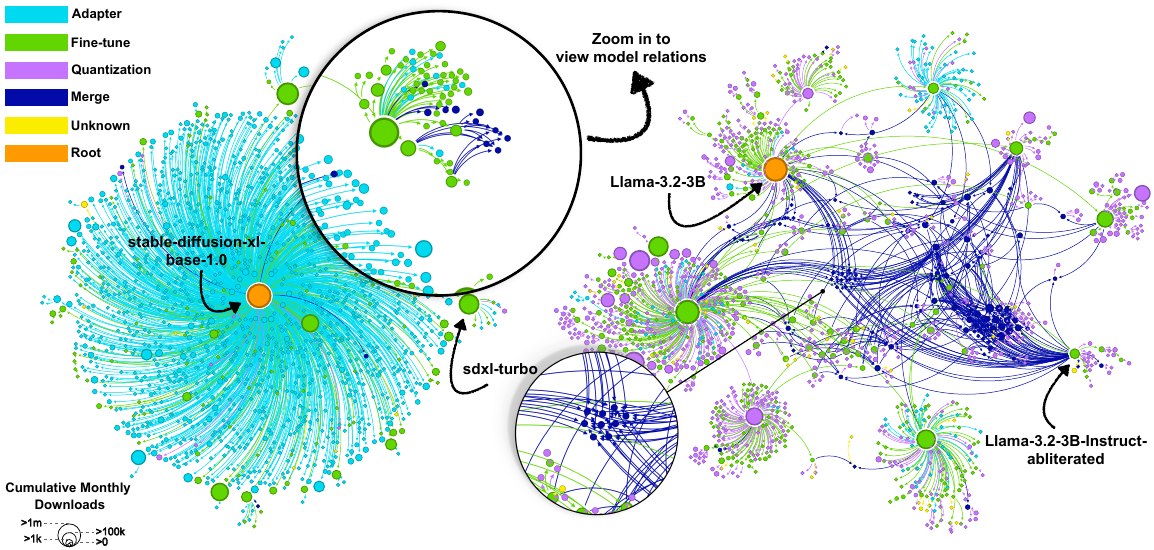}
\caption{\textit{\textbf{The Model Atlas - Stable Diffusion vs. Llama:}} We visualize the atlas of the top 30\% most downloaded models in the Stable Diffusion (SD) and Llama regions. Node size reflects cumulative monthly downloads, and color denotes the transformation type relative to the parent model. The atlas reveals that the Llama region has a more complex structure and a wider diversity of transformation techniques (e.g., quantization, merging) compared to SD. \textit{Zoom in to view edges, best viewed in color.}}

 \label{fig:llama_vs_sd_atlas}
\end{figure}

Model forensics allows us to trace a model’s training trajectory, its origins, the data used, and the transformations it underwent \citep{spectral_detuning,mother,dsire,phylolm,neural_phylogeny,neural_lineage,carlini2024stealing}. This has major implications for intellectual property, bias estimation, safety, and reproducibility. The atlas also enables intuitive visualizations of huge model populations, facilitating \textit{\q{meta-ML research}} to analyze the evolving ML landscape. Concretely, we demonstrate that the atlas can reveal structural patterns, emerging trends, and opportunities to transfer knowledge between communities. For instance, in \cref{fig:llama_vs_sd_atlas} we present a preliminary atlas which reveals distinct structural patterns across modalities, Llama-based models \citep{dubey2024llama} have more diverse and complex training dynamics (e.g., quantization and model merging) than Stable Diffusion \citep{stable_diffusion}. Finally, with so many models available, we need tools for accurate model discovery \citep{lu2023content,li2021ranking,ding2022pactran,huang2022frustratingly,you2021logme,nguyen2020leep,tran2019transferability,zhang2023model} which could reduce training costs, shorten development cycles, and lower environmental impact. Model discovery requires knowing what each model does and how well it performs, information that the atlas captures. While related to prior work (e.g., \citet{model_lakes}), our position focuses specifically on the Model Atlas and the methods required to chart it.

Although many models exist, most are undocumented and contain little to no usable metadata. Over 60\% of models on HF have no documentation at all \citep{mother,probex,probelog}, and for the remainder, much is incomplete. This includes many of the platform’s most popular models, which often omit key details such as training data, performance, or lineage. Moreover, public repositories represent only a fraction of the global model landscape; private repositories in companies and research labs contain many more models. Consequently, the Model Atlas remains mostly uncharted, populated by \q{lost models}, whose  origins, capabilities, and interrelations are largely unknown. Completing the atlas requires inferring its missing nodes, edges, and their attributes. This motivates a new class of ML problems that treat models themselves as data. However, learning directly from models presents major challenges: weights are high-dimensional, exhibit harmful permutation symmetries, and lack established structural priors. To address this, the emerging field of weight-space learning \citep{model_lakes,nws,schurholt2021self,schurholt2022hyper,weight_space_workshop} develops specialized neural networks that take other models as input and infer their properties. The dominant approach involves designing \textit{equivariant networks}, which account for the permutation symmetries of model weights. While promising, current methods remain computationally expensive and are mostly limited to small-scale models or simplified settings.

We argue that scaling atlas charting to large, diverse model populations requires rethinking the role of weight symmetries. Instead of tackling permutation invariance directly, we propose a path that sidesteps it altogether. By avoiding symmetry constraints, we can leverage standard architectures and training recipes, enabling more efficient and scalable methods. We highlight concrete scenarios where this is already feasible, such as probing models for their functional responses or learning on weights within symmetry-consistent subsets. While these strategies are still in their early stages, they demonstrate the practicality of symmetry-agnostic approaches. Realizing the full potential of the Model Atlas will require new research into priors, architectures, and learning methods. We believe this is both a timely and worthwhile effort that would benefit broad community participation.

\section{The Model Atlas}
\label{sec:definitions}
We aim to represent model collections in a way that captures each model’s lineage, creation process, functional capabilities, performance, and metadata. We therefore propose the \emph{Model Atlas}, a directed acyclic graph (DAG) denoted by $\mathcal{G} = (\mathcal{V}, \mathcal{E})$, where each node $v \in \mathcal{V}$ represents a model at a specific point in its training trajectory. Directed edges $(u, v) \in \mathcal{E}$ denote a weight transformation from model $u$ to model $v$, e.g., fine-tuning or quantization.

\begin{wrapfigure}[16]{R}{0.65\textwidth}
\centering
\includegraphics[width=0.65\textwidth]{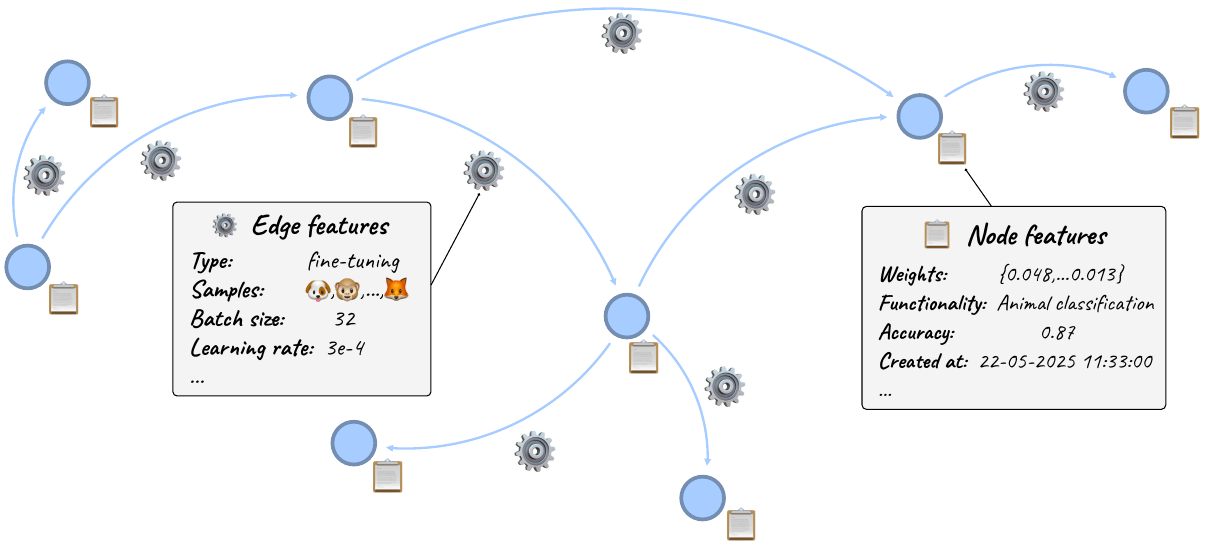}
\caption{\textit{\textbf{Model Atlas illustration:}} Each node in the Model Atlas represents a distinct model state, and each directed edge denotes a weight transformation from one model to another. Edge features encode information about the transformation, node features capture properties of the model itself.}
\label{fig:model_atlas_example}
\end{wrapfigure}

Both nodes and edges have associated features. Node features include all the information about a model, e.g., its weights, functional traits, and metadata. Traits capture the functional and behavioral properties of a model, such as accuracy, robustness, and fairness. Unlike traits, which reflect learned behaviors, metadata captures static or descriptive attributes such as creation time and license type. Edge features capture information about the transformation from parent model $u$ to child model $v$, e.g., optimization parameters, training data, and transformation type (e.g., LoRA \citep{lora} or quantization).

This graph contains virtually all information about the model population, including how each model was created, its relation to other models, and its capabilities and metadata (see \cref{fig:model_atlas_example} for an overview). The atlas provides the basis for a wide range of applications that we describe in \cref{sec:application}. However, as we describe in \cref{sec:methods}, the core obstacle is that only a small part of this graph is known; for most nodes, only a very small portion of the details about their edges, node features, or edge features are given. To unlock the rich rewards of the Model Atlas, we believe it is paramount to develop new ML tools for recovering the rest of the graph.

\section{Why is the Model Atlas useful?}
\label{sec:application}
The Model Atlas is central to our position, as it provides a structured way to represent model collections. Its utility stems from the many applications that it enables. We group these into three main categories: model forensics, meta-ML research, and model discovery. Note, some applications rely on access to the atlas's structure and contents, which are mostly unknown in practice. In \cref{sec:methods} we describe scalable methods for charting the missing regions of the atlas. 

\subsection{Model forensics} 

Neural networks are complex, opaque functions. They do not readily reveal details about their training data or predecessor models. However, this information is critical, particularly for the creators of training datasets or upstream models. By charting models within the atlas, we can infer such provenance relationships. The atlas facilitates this in several ways. First, a directed path between two models indicates which model is descended from the other. Second, if a model descends from others, it implicitly inherits their training data. Since edge features include information about the data used during transformation, the atlas can reveal the full data lineage of a model, an important capability for intellectual property and compliance. The same ideas also extend to bias estimation. If an ancestor exhibits a known bias, we can propagate that knowledge downstream. Finally, the atlas improves reproducibility by recording the exact parameters used to transform one model into another.

\begin{figure*}[t]
    \centering
    \includegraphics[width=\linewidth]{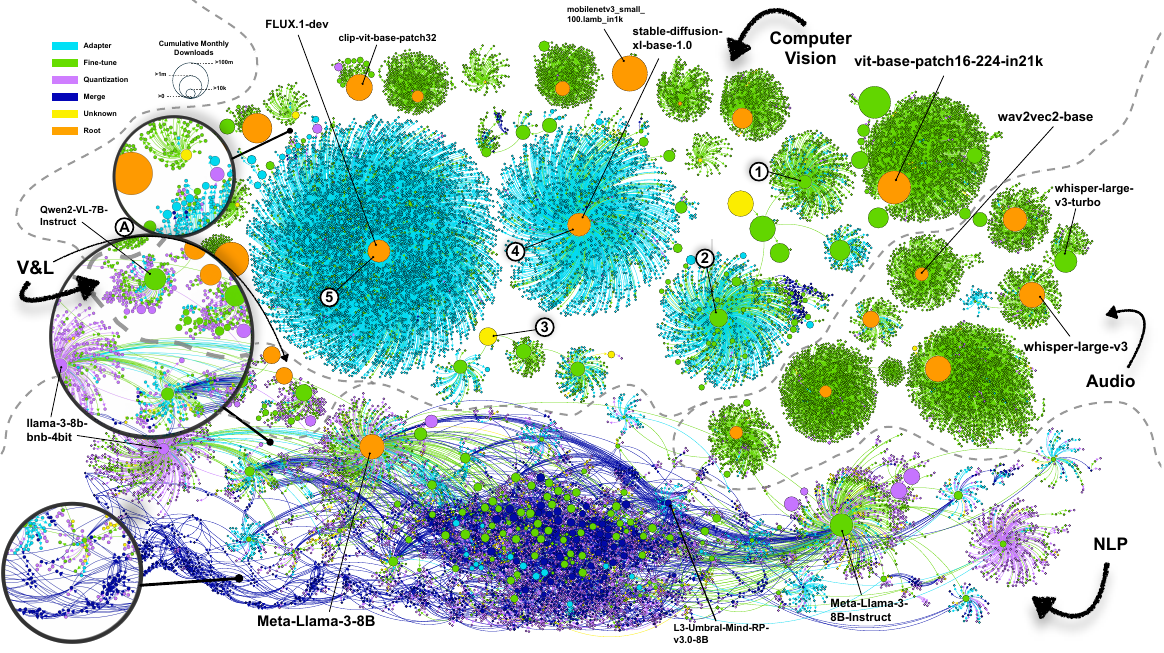} 
\caption{\textit{\textbf{The Hugging Face Model Atlas:}} While this is a small subset (63,000 models) of the documented regions of Hugging Face, it already reveals significant trends.  \textit{Depth and structure.} The LLM connected component (CC) is deep and complex. It includes almost a third of all models. In contrast, while Flux is also substantial, its structure is much simpler and more uniform. \textit{Quantization.} Zoom-in (A) highlights quantization practices across vision, language, and vision-language (V\&L) models. Vision models barely use quantization, despite Flux containing more parameters (12B) than Llama (8B). Conversely, quantization is commonplace in LLMs, constituting a large proportion of models. VLMs demonstrate a balance between these extremes. \textit{Adapter and fine-tuning strategies.} A notable distinction exists between discriminative (top) and generative (bottom) vision models. Discriminative models primarily employ full fine-tuning, while generative models have widely adopted adapters like LoRA. The evolution of adapter adoption over time is evident: Stable-Diffusion 1.4 (SD) (1) mostly used full fine-tuning, while SD 1.5 (2), SD 2 (3), SD XL (4), and Flux (5) progressively use more adapters. Interestingly, the atlas reveals that audio models rarely use adapters, suggesting gaps and opportunities in cross-community knowledge transfer. This inter-community variation is particularly evident in \textit{model merging}. LLMs have embraced model merging, with merged models frequently exceeding the popularity of their parents. This raises interesting questions about the limited role of merging in vision models. For enhanced visualization, we display the top 30\% most downloaded models. \textbf{The image is high-resolution, zoom in to view individual nodes and edges. Best viewed in color.}}
    \label{fig:motivation}
\end{figure*}

\subsection{Meta-ML research by visualizing entire model collections} 
The Model Atlas enables large-scale visualization and analysis of model repositories, a capability we refer to as \textit{\q{meta-ML research}}. By encoding node and edge features into visual attributes, such as color and size, these visualizations\footnote{We use \q{Gephi} by \citet{bastian2009gephi} to visualize the Model Atlas.} provide an interpretable overview of structure, model attributes, and emerging trends across vast model populations. Note that node position is optimized for clarity \citep{jacomy2014forceatlas2} and does directly reflect distance between model weights. To demonstrate this, we analyze Hugging Face (HF), the largest public model repository, which hosts over 1.5 million models, with more than 100,000 added each month. While HF encourages documentation via Model Cards \citep{model_cards}, over 50\% of models lack them entirely. Additionally, fewer than 30\% of models specify their parent model. Using these, we construct an \textit{initial} atlas and explore recent trends. In \cref{fig:motivation}, we visualize a portion of this atlas, comprising over 60,000 models across 28 connected components and more than 65,000 edges. We use this structure to compare development patterns in computer vision (CV) and natural language processing (NLP). See \cref{app:additional_meta_ml} for further visualizations and analysis. We will make the Model Atlas used for these visualizations publicly available through an interactive web interface that allows users to navigate and explore the different models.

\noindent \emph{Quantization.} As shown in \cref{fig:motivation}, quantization is rare in CV models (fewer than $0.15\%$ of all models in this pool) compared to NLP. This suggests that vision models have not yet reached the scale where inference cost necessitates quantization. However, using the atlas we can spot an emerging change of trend. The presence of a highly downloaded quantized version of Flux \cite{flux2024}, one of the largest generative vision models, indicates that image generation models may have just reached the scale where quantization is valuable (see zoom-in (A) in \cref{fig:motivation}).

\noindent \emph{Fine-tuning vs. adapters.} Generative and discriminative models exhibit markedly different adaptation strategies. Most generative models rely on parameter-efficient adapters (e.g., LoRA), whereas discriminative models almost exclusively use full fine-tuning (see \cref{fig:motivation}-top). The Model Atlas not only highlights this distinction but also reveals how these trends are shifting over time. In earlier generations, such as SD1.4, only around $50\%$ of models used adapters. In contrast, newer generative models like Flux \citep{flux2024} and Llama 3 \citep{dubey2024llama} overwhelmingly adopt adapter-based methods, indicating a broader move toward more efficient fine-tuning (see pins (1)-(5) in \cref{fig:motivation}).

\noindent \emph{Merging.} Model merging \citep{ties_merging,shah2023ziplora,wortsman2022model,stoica2024model,yadav2024survey} is about 35 times more common in NLP than in CV. On average, merged NLP models achieve 30\% higher influence (measured via descendant downloads; see \cref{app:additional_meta_ml}) than their non-merged siblings. While this does not imply causation, it suggests that merging is an underexplored strategy in vision.

\subsection{Model discovery} 
Training new models is time-consuming, expensive, and energy-intensive. Despite this, practitioners often train or fine-tune a new model for each task. Imagine that instead, one could retrieve a suitable pre-trained model and use it directly. This would reduce costs, shorten development cycles, and lower environmental impact. The atlas provides the infrastructure needed for such model discovery. It enables search based on model capabilities, performance, or transformation history. This goes far beyond search functionality in existing repositories, which mostly rely on textual search of sparse documentation. For example, although tens of thousands of models are trained on ImageNet \citep{imagenet}, searching for the class \q{peacock} among the $< 1.5M$ HF models returns fewer than $100$ results. In contrast, an atlas-based search, leveraging traits and lineage, would yield many more relevant results.

If no existing model fits a given task, the Model Atlas can still assist by helping identify models, or sets of models, likely to transfer well. Estimating transferability \citep{lu2023content,li2021ranking,ding2022pactran,huang2022frustratingly,you2021logme,nguyen2020leep,tran2019transferability,zhang2023model} is computationally expensive, as existing methods typically require evaluating each model on the target dataset. In contrast, a complete atlas enables graph-based search strategies that reduce the number of models that must be tested. For example, a branch-and-bound strategy could prune subgraphs unlikely to yield high transfer performance. Similarly, multi-scale search techniques, akin to binary search, could accelerate retrieval by narrowing the candidate space efficiently. We believe that a complete Model Atlas will enable scalable transferability estimation and more effective model reuse.

\section{The atlas is incomplete, we need new methods to chart it}
\label{sec:methods}

\begin{wrapfigure}[14]{R}{0.3\textwidth}
\centering
\includegraphics[width=0.24\textwidth]{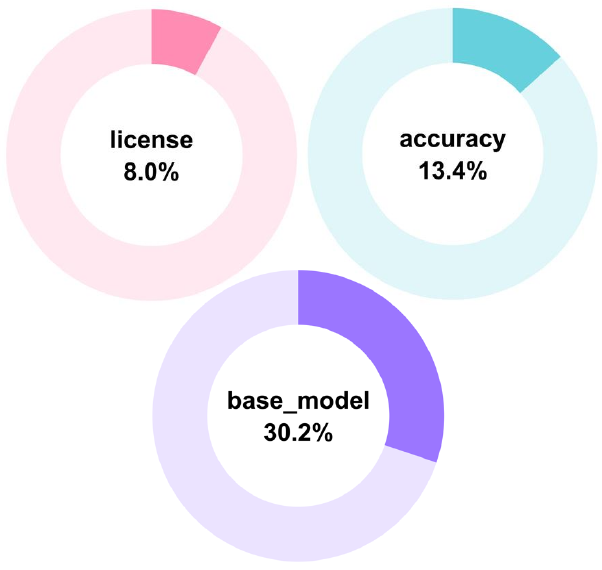}
\caption{\textit{\textbf{Documentation levels in HF:}} Most models on Hugging Face suffer from poor documentation quality.}
\label{fig:documentation_rings}
\end{wrapfigure}

\subsection{The atlas is incomplete}
Despite the abundance of models, most are undocumented, leaving little usable information. To address this, solutions like Model Cards \citep{model_cards} were introduced to provide standardized descriptions. While an important first step, current documentation practices fall short of handling the scale and complexity of modern model ecosystems. As highlighted earlier, over 60\% of models on HF lack \textit{any} model card. Worse still, even documented models typically provide only sparse or inconsistent information. For instance, fewer than 15\% of HF models report accuracy, and only 8\% include license details (see \cref{fig:documentation_rings}). Since this data is self-reported, its quality varies widely. Consequently, recent attempts to summarize documentation automatically (e.g., by HF \citep{smolhubtldr}) face fundamental limitations: missing information cannot be recovered through summarization alone. Overall, current practices lack the required coverage, consistency, and adaptability.

This documentation failure has created a vast population of \q{lost models}, models whose origins, capabilities, and relationships are largely unknown. This has tangible consequences. Organizations routinely waste resources re-training models that already exist but are undiscoverable. Moreover, even failed models contain valuable insights into architecture, data, and optimization; this implicit knowledge is currently lost, forcing the community to rediscover it. The atlas helps preserve this implicit knowledge and serves as a tool for knowledge retention.

\subsection{Charting the Model Atlas}
Although we advocate the Model Atlas as a foundational structure, much of it remains missing. Completing the atlas requires inferring its missing nodes, edges, and their attributes. This motivates a new class of ML problems where models (or sets of models) serve as input, and the goal is to predict properties of the atlas. While ML for images, text, and audio is well-established, learning directly from model weights is only starting. This presents two key challenges. First, model weights are high-dimensional, making them computationally difficult to process. Second, neural network weights exhibit parameter space symmetries \citep{hecht1990algebraic}. For instance, permuting neurons in hidden layers of an MLP does not affect the model’s function, but does change the raw weight representation. This property complicates the development of techniques that generalize across diverse model populations, as most ML methods do not naturally account for these symmetries.

\begin{wraptable}[28]{R}{5.5cm} 
    \vspace{-1\baselineskip} 
    \small 
    \centering 
    \caption{\textit{\textbf{Atlas-based documentation imputation:}} Using atlas structure improves prediction of model accuracy and other attributes, compared to naively using the majority label. In (b), we report the prediction accuracy.} 
    \label{tab:main_imputation} 

    \begin{subtable}{\linewidth} 
        \centering 
        \setlength{\tabcolsep}{3pt} 
        \caption{Metric Imputation Results} 
        \label{tab:metric_imputation} 
        \begin{tabular}{@{}llccc@{}} 
             & \textbf{Method} & \textbf{MSE} & \textbf{MAE} & \textbf{Corr.} \\ 
            \midrule
            \multirow{5}{*}{\rotatebox{90}{\begin{tabular}[c]{@{}c@{}}TruthfulQA \\ (0-shot, mc2)\end{tabular}}}
            & Baseline   & 100.217 &  8.541 & - \\
            & Ours 1-NN  & 32.830  & 3.247  & 0.856 \\
            & Ours 2-NN  & 28.720  & 3.235  & 0.864 \\
            & Ours 3-NN  & 25.544  & 3.147  & 0.877 \\
            & Ours 5-NN  & 23.512  & 3.093  & 0.885 \\
            \midrule
            \multirow{5}{*}{\rotatebox{90}{\begin{tabular}[c]{@{}c@{}}Helleswag \\ (0-shot)\end{tabular}}}
            & Baseline   & 95.000  & 7.500  & - \\
            & Ours 1-NN  & 30.000  & 3.000  & 0.860 \\
            & Ours 2-NN  & 27.000  & 2.900  & 0.870 \\
            & Ours 3-NN  & 24.000  & 2.800  & 0.880 \\
            & Ours 5-NN  & 22.000  & 2.700  & 0.890 \\
        \end{tabular}
    \end{subtable}

    \vspace{\medskipamount} 

    \begin{subtable}{\linewidth} 
        \centering 
        \setlength{\tabcolsep}{3pt} 
        \caption{Attribute Prediction Accuracy} 
        \label{tab:hubs_imputation} 
        \begin{tabular}{@{}lcc@{}} 
            \textbf{Attribute} & \textbf{Graph Avg.} & \textbf{Hub Avg.} \\
            \midrule
            \texttt{pipeline\_tag} & 0.60  & 0.79  \improve{0.19} \\
            \texttt{library\_name} & 0.81  & 0.84  \improve{0.02} \\
            \texttt{model\_type} & 0.66  & 0.81  \improve{0.15} \\
            \texttt{license} & 0.49  & 0.85  \improve{0.35} \\
            \texttt{relation\_type} & 0.61  & 0.80  \improve{0.19} \\
            
        \end{tabular}
    \end{subtable}

\end{wraptable}

The emerging field of studying models based on their weights is called Weight-Space Learning \citep{statnn,nws,schurholt2021self,schurholt2022hyper,weight_space_workshop}. Research in this area generally falls into $3$ categories: i) \textit{Learning weight representations}, recovering model functionality and performance \citep{probegen,probex,probelog,dsire,schmidhuber,lim2023graph,navon2023equivariant_alignment,navon2023equivariant,kofinas2024graph,neural_functional_transformers,lol}. ii) \textit{Developing generative models for weights}, synthesizing of new model parameters \citep{ha2016hypernetworks,nern,peebles2022learning,erkocc2023hyperdiffusion,weights2weights,sane,ties_merging}. And iii) \textit{Recovering weight trajectories}, tracing the evolution of model parameters over time \citep{spectral_detuning,mother,chen2022copy,carlini2024stealing,phylolm,neural_lineage,neural_phylogeny,xu2024instructional,yang2024fingerprint}. Despite promising progress, most work in this space is limited to synthetic or simplified settings. Current approaches cannot yet handle many of the subtasks and scale involved in charting real-world model populations.

\emph{Avoiding, not embracing equivariance.} The dominant approach in weight-space learning is to design equivariant networks \citep{lim2023graph,navon2023equivariant_alignment,navon2023equivariant,kofinas2024graph,neural_functional_transformers,zhou2024universal,zhou2024permutation,lol,shamsian2024improved,lim2024empirical} that respect the permutation symmetries of neural weights. This is an elegant but computationally costly endeavor, requiring specialized adaptations for different layer types (e.g., convolutional, self-attention, or state-space). Unfortunately, standard deep learning architectures are not equivariant by default, and current solutions remain too expensive for web-scale model repositories. Rather than redeveloping architecture classes from scratch, we argue for an alternative path: reformulating tasks to avoid permutation symmetry altogether. This allows us to use standard architectures and training tricks. While not all weight-space learning tasks admit such reformulations, many graph-based tasks do.

\emph{Graph kNN.} One simple approach is to predict properties of unlabeled models based on their neighbors in the partially charted atlas. For example, we used the Mistral-7B connected component (17.5k models, with only 300 labeled on the \textit{TruthfulQA (0 shot) (mc2)} metric; see \cref{app:dataset_details} for more details). We estimated performance for each unlabeled node by averaging the scores of its $k$ nearest neighbors, where distance is defined by path length in the undirected atlas. \cref{tab:metric_imputation} shows that this simple method can predict model accuracy surprisingly well. Beyond accuracy, the atlas can also help infer missing metadata. We use the concept of \textit{model hubs}, defined as sets of sibling leaf nodes ($79\%$ of models belong to some hub). Assuming models in the same hub are similar, we impute missing attributes using majority voting across labeled nodes. We tested hub-based predictions on five attributes that are often missing on HF, including: license, model and inheritance types (see \cref{fig:documentation_rings}). Compared to global majority baselines, hub-based prediction significantly improves accuracy, by 35\% for license prediction and 19\% for both inheritance type and pipeline tag (\cref{tab:hubs_imputation}). These methods are simple and scalable, though not always sufficiently precise and limited by the completeness of the known atlas.

\emph{Learning on functional features.} Permutation symmetry affects the parameter space, not the function space. One natural workaround is to represent a model via its responses to predefined inputs. Probing-based methods \citep{probelog,probegen,schmidhuber} select a set of $m$ probes (inputs) $x_1, \dots, x_m$ and represent the model as the concatenation of its responses:

\begin{equation*}
   \phi(f) = (f(x_1),f(x_2),\dots,f(x_m))
\end{equation*}

\begin{figure*}[t]
    \centering
    \includegraphics[width=\linewidth]{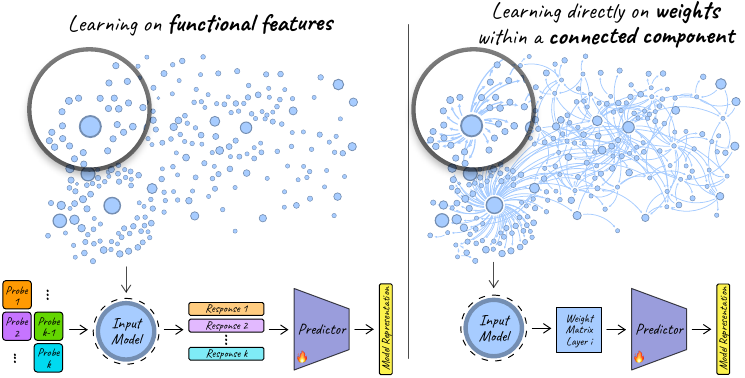} 
\caption{\textit{\textbf{Equivariance-bypassing strategies:}} When working with a general collection of models, one can probe each model and train a model on its responses, thereby sidestepping the challenge of weight permutations (left). In contrast, within a connected component of the atlas, neuron permutations are mostly fixed, allowing us to train standard architectures directly on the weight matrices of individual layers (right). This significantly reduces computational cost. In both cases, these strategies support scalable charting of large, in-the-wild populations. The fire symbol indicates the learned component.}
    \label{fig:tree_vs_probing}
\end{figure*}

As long as the responses exhibit structure, we can learn downstream tasks using standard architectures (see \cref{fig:tree_vs_probing}-left). Similarly to a good examiner, the key is to know which questions to ask. While determining the optimal set of probes is an exciting, outstanding challenge, \citet{probegen} showed that even quite native probes performed better than many state-of-the-art methods. The main limitation of probing methods is their runtime, as computing their representations requires $m$ forward passes per model, which can become costly at scale.

\emph{Learning directly on weights within a connected component.} A core motivation of equivariant architectures is that weights suffer from permutation symmetries. But it turns out that this is not always the case. A child model keeps mostly the same neuron ordering as its parent model, as finetuning only changes weights very slightly. Recursively, all models fine-tuned from a foundation model will typically have the same permutation. More formally, all models within the same connected component in the Model Atlas have consistent neuron ordering. In such cases, we can apply standard architectures directly to weights without equivariance (see \cref{fig:tree_vs_probing}-right). The outstanding challenge is to develop architectures able to deal with the extremely high dimension of network weights (millions to billions). While current architectures \cite{probex} use factorized classifiers, taking in a single model layer, we believe better architectures, that span multiple layers, will be better.  

Working within the same connected component allows for simpler tools, as \citep{mother} demonstrated, computing a node's nearest neighbors provides a clue to what nodes are connected to it by an edge, allowing link prediction. Clustering models by their weights also reveals other shared attributes \citep{gueta2023knowledge,mother}. We extend this insight by introducing a link prediction process that incorporates real-world structural priors extracted with the help of meta-ML research using visualization of HF. By analyzing patterns such as quantization, duplication, temporal upload dynamics, and checkpoint structures, we develop simple yet effective decision rules that guide edge prediction. These priors allow us to move beyond generic tree-based heuristics and chart realistic DAGs while significantly improving the accuracy at a lower computational cost. See \cref{sec:charting_method} for full implementation and results.

\section{Open challenges in Model Atlas charting}
\label{sec:challenges}
Charting the full atlas will require new research into priors, architectures, and learning methods. We now outline key open challenges; see \cref{app:additional_open_challenges} for further discussion.

\noindent \emph{New visualization methods.} In this paper, we visualized the Model Atlas using off-the-shelf graph layout algorithms \citep{jacomy2014forceatlas2,bastian2009gephi}. While these are already useful for meta-ML research, they are not designed for the unique structural and semantic characteristics of Model Atlases. Future visualization methods could exploit specific structural priors, such as large hubs and near tree-like hierarchies. Additionally, incorporating node and edge features (e.g., model metadata, weight distance metrics, or performance estimates) could help position semantically related models closer together. Moreover, most existing graph layouts also assume static graphs, whereas Model Atlases evolve continuously. Supporting dynamic insertions of nodes and edges will require new online layout algorithms.

\noindent \emph{Charting priors and weight-space learning architectures.} The approaches highlighted in \cref{sec:application} exploit specific patterns and priors that allow them to bypass the challenge of weight permutation. However, given the scale of modern model repositories and the size of individual models, extending charting efforts to the web scale will likely require new priors and architectures. In particular, future work may explore downsampling strategies that retain essential semantic information, enabling efficient processing of large model populations. Most current methods also assume access to model weights. Extending charting to black-box settings, such as proprietary APIs like ChatGPT or Gemini \citep{chatgpt,gemini}, will require new techniques based on outputs, activations, or metadata.

\noindent \emph{Recovering training trajectories.} Most current weight-space learning methods focus on node-level properties such as function or performance. However, recovering edge-level attributes, e.g., optimizer, data, or learning dynamics, is equally important. Existing methods often assume single-parent transitions and do not account for more complex scenarios such as model merging or distillation. Extending charting algorithms to account for multi-parent edges, distillation relationships, and non-weight-based transformations remains a largely unexplored challenge. More broadly, current atlases can be viewed as sparse samples of a larger underlying atlas capturing the continuous training trajectory. In most cases, edges in the atlas represent transitions between model states that may have occurred through multiple intermediate optimization steps. Recovering this full, fine-grained Model Atlas, including the intermediate state, presents a compelling and open challenge for future research.

\noindent \emph{Probe selection methods.} Probing-based approaches provide a powerful way to represent models functionally, but they involve a tradeoff: using too many probes increases runtime, while too few increases sensitivity to probe choice. To overcome this, we need new methods for selecting a minimal yet expressive set of probes. This could involve real input probes (e.g., curated data points), requiring principled selection strategies, or synthetic probes, which call for generative or optimization-based techniques. Designing robust, task-adaptive probing schemes is an open problem.

\section{Alternative view}
\label{sec:alternative_view}
While we strongly advocate for charting model collections through the Model Atlas, it is important to acknowledge valid alternative perspectives.

\noindent \emph{Foundation models may render model collections obsolete.} One view is that the rise of increasingly capable foundation models, such as ChatGPT and Gemini \citep{chatgpt,gemini}, may reduce the relevance of smaller, task-specific models. If this consolidation continues, some might argue that model repositories will shrink or become unnecessary, thereby reducing the importance of charting them. While reasonable, this perspective does not align with current trends. In practice, the number of public models continues to grow. Moreover, even a small number of dominant models would produce numerous variants, checkpoints, and fine-tuned descendants. These models would still benefit from organization, provenance tracking, and comparison, tasks the atlas is ideal for.

\noindent \emph{Equivariant methods may eventually scale.} Our position is that equivariant methods will struggle to scale to full repositories. A valid alternative view is that given sufficient research effort, the field will find the right architecture, training tricks, and hyperparameters to make them efficient alternatives. While this is a different technical approach from the one we advocate, this would not change the formulation and applications of the atlas, which is the essential part of our position. Furthermore, this is unlikely to help within connected components where there are no permutation symmetries.

\noindent \emph{Charting may not require machine learning.} A valid argument is that charting the Model Atlas could be addressed by simply tightening documentation requirements, eliminating the need for new machine learning methods. Under this approach, model creators would be required to upload all relevant information alongside the model or embed it directly within the weights file. While such protocols may eventually be adopted, they are unlikely to capture all the information we might want about a model. They would also have no effect on the millions of existing models that remain undocumented. Finally, we draw some parallels with code documentation, which is good practice and encouraged by all, but is rarely done. We believe forcing model creators to document is doomed to fail.

\section{Conclusion}
\label{sec:conclusion}
The rapid growth in the number of trained models presents an opportunity to study them collectively rather than in isolation. The Model Atlas formalizes this perspective by representing models, their attributes, and the transformations that link them in a unified structure. We have outlined its potential applications in model forensics, meta-ML research, and model discovery. The central challenge is that most edges in the graph, as well as the majority of node and edge features, are missing. Addressing this gap calls for a new class of machine learning methods that treat neural networks as data and infer meaningful properties from them. We argue that the most promising approaches are those that sidestep the symmetries inherent in weight space, enabling scalable solutions across diverse model populations. We hope that the Model Atlas will inspire broader research into understanding and organizing the global population of models.

\newpage

{
    \small
    \bibliographystyle{ieeenat_fullname}
    \bibliography{main}
}

\newpage


\appendix

\begin{figure}[b]
\centering
\begin{subfigure}[b]{0.485\textwidth}
    \centering
    \includegraphics[width=0.9\linewidth]{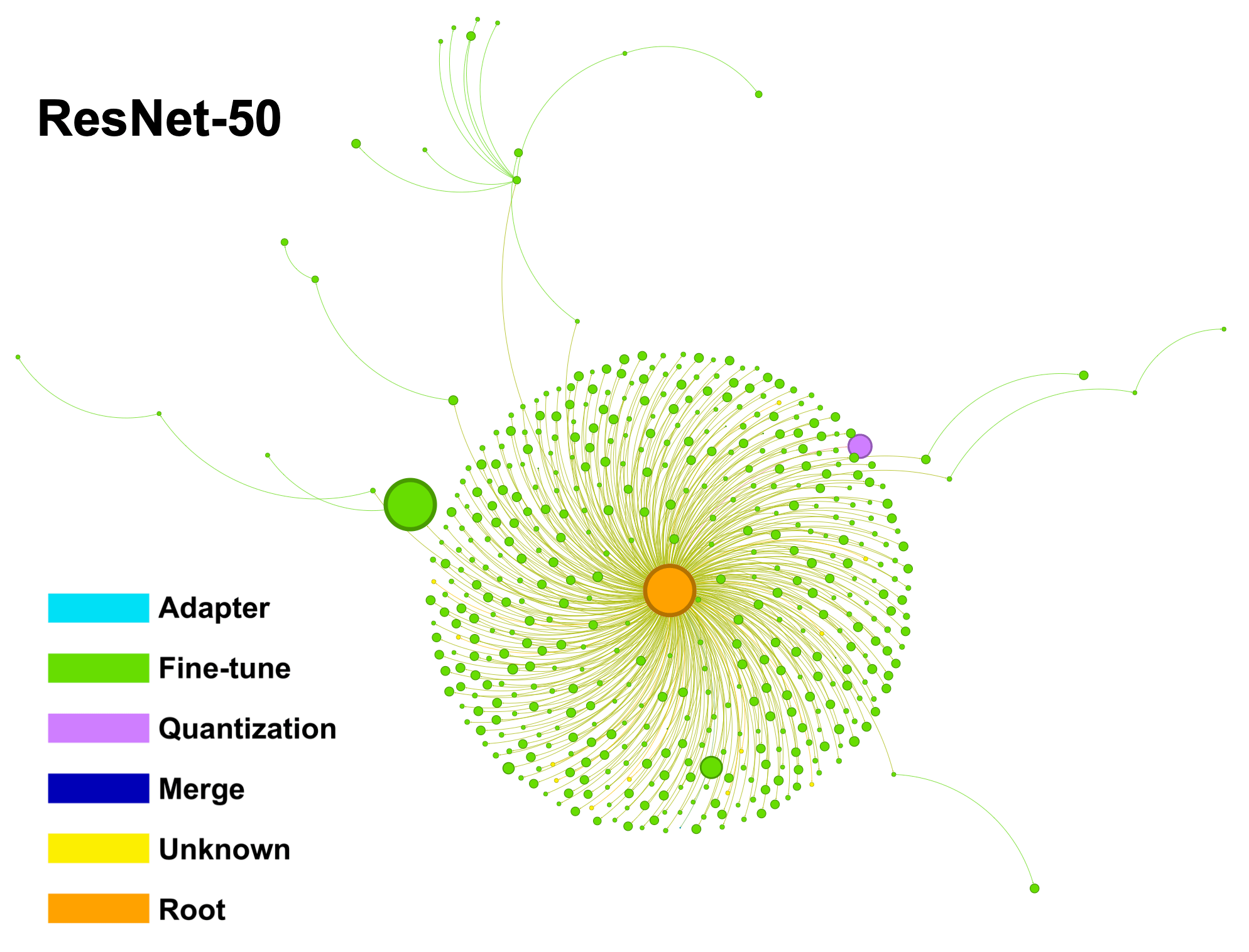}  \\   
    \caption{\textit{\textbf{Individual connected components - 1/6}}}
    \label{fig:cc1}
\end{subfigure}
\hfill
\begin{subfigure}[b]{0.485\textwidth}
    \centering
    \includegraphics[width=0.9\linewidth]{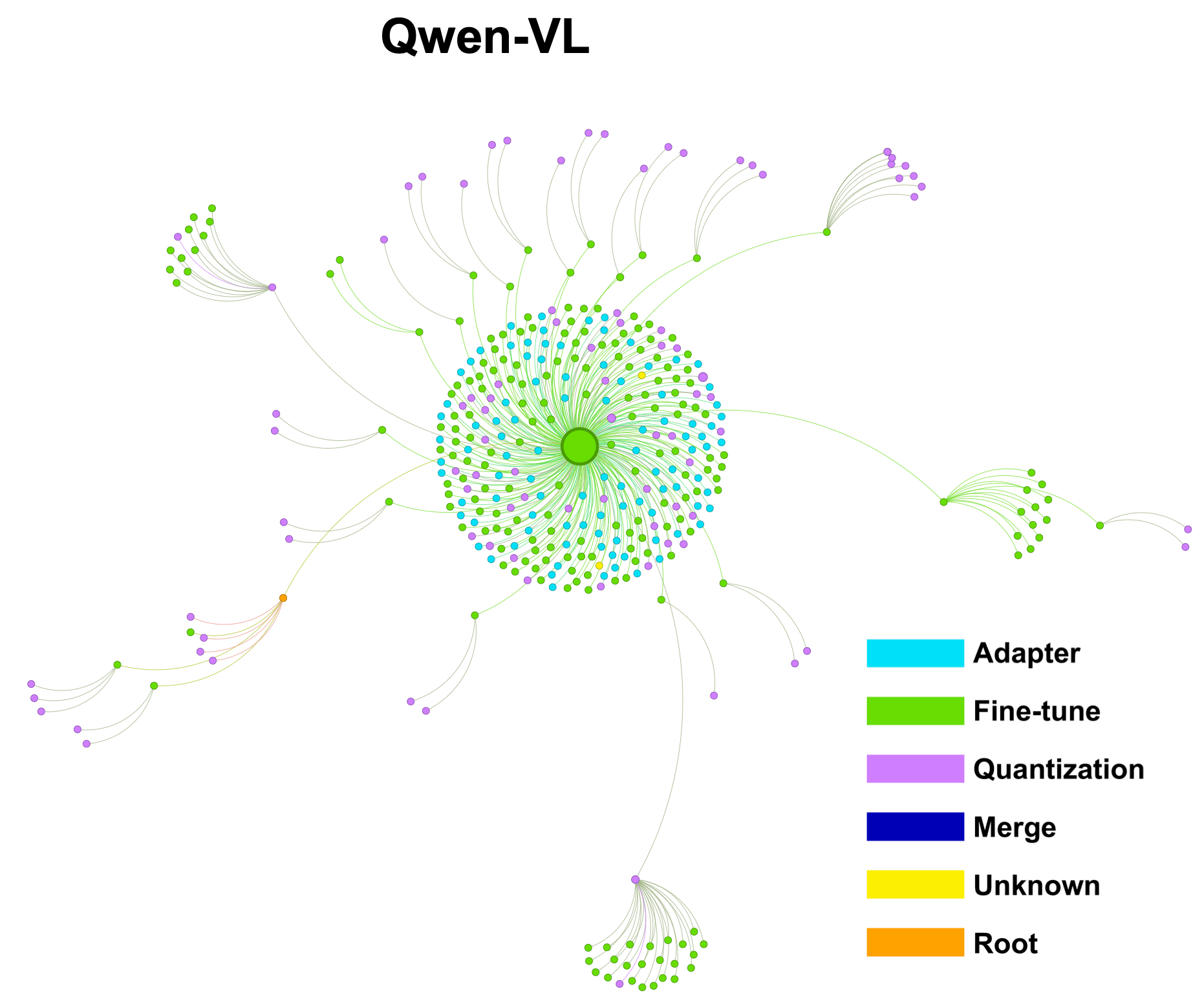}  \\  
    \caption{\textit{\textbf{Individual connected components - 2/6}}}
    \label{fig:cc2}
\end{subfigure}
\end{figure}

\section{Additional meta-ML research insights and visualizations}
\label{app:additional_meta_ml}
\begin{wrapfigure}[29]{R}{0.3\textwidth}
\centering
\includegraphics[width=0.26\textwidth]{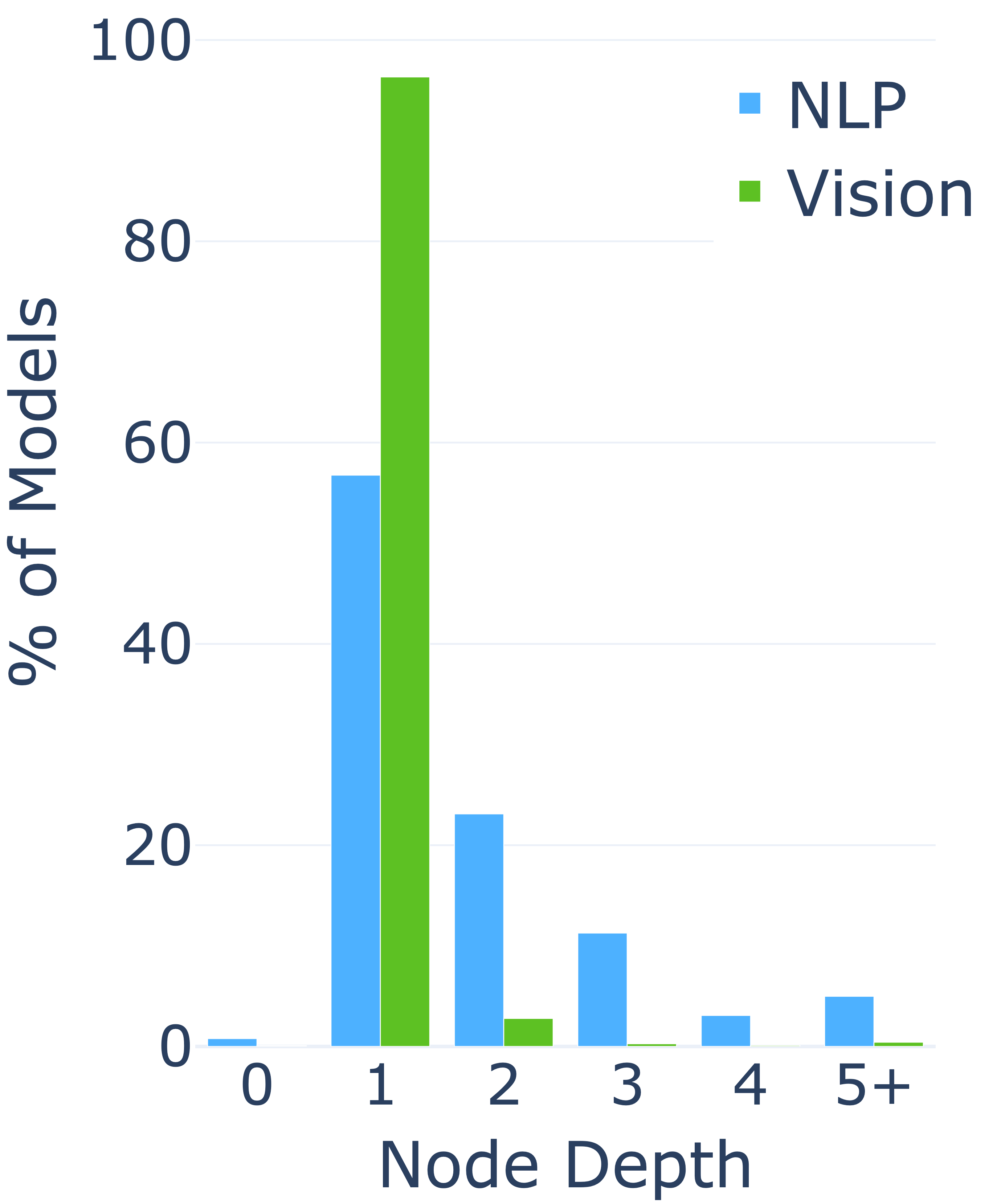}
\caption{\textit{\textbf{Vision vs. NLP Node depth:}} By analyzing over 314k models, we found that over $96\%$ of CV models are situated one node away from the root, while only $55\%$ of NLP models have this shallow depth. Over $5\%$ of NLP models have depth of at least five nodes. This shows that NLP models are much deeper than CV models, suggesting the NLP community embraces iterative refinement over moving to the latest foundation models.}
\label{fig:model_depth}
\end{wrapfigure}

\noindent \emph{Depth.} \cref{fig:model_depth} shows that NLP models have a wide range of DAG depths, while in CV, nearly all models have a direct edge to the root foundation model. This suggests that the CV community puts more focus on new foundation models, while the NLP community often embraces iterative refinement. As an example, in \cref{fig:motivation} the LLM connected component (CC) exhibits significant depth and complexity, representing almost a third of the models. In contrast, while Flux is also substantial, its structure is much simpler and more uniform.

\noindent \emph{Better model impact measurement.} 
There are several ways of measuring the impact of a model. For example, HF uses likes, trends (based on an undisclosed proprietary algorithm), and downloads. These metrics are somewhat myopic, as they measure the direct popularity of models but not the popularity of their descendant models. In fact, our graph analysis reveals that for $50\%$ of non-leaf nodes, the total downloads of their descendants exceed their own individual downloads. This is partially due to the popularity of quantized child models and to incremental improvement of child models, e.g., finetuning or merging. Our analysis further showed that for non-leaf models, the sum of descendant nodes downloads exceeds those of the model itself in most cases (often by large margins). This suggests that simple model download counts underestimate the influence of the parent model.

We can therefore use the atlas to introduce a new model impact metric: \verb|sub_tree_downloads|.  We calculate this metric by summing the downloads of the model node and those of all its descendants. This number describes how many downloads this model causally affected. It has important applications to intellectual property rights, as the models and data used to train the target models affected all of its downstream downloads. It also quantifies the social impact of the biases of this model.

\noindent \emph{Restoring removed models.} There are legal and commercial reasons for removing models from repos. Some models have been removed from the repository over time, affecting the integrity of the model DAGs. Out of $33,870$ identified source models, $1,612$ are missing due to deletion. A notable case is \textit{runwayml/stable-diffusion-v1-5}, which has $3,038$ broken references. To preserve the integrity of the dependency structure, the missing node was reconstructed and its connections manually restored. Using methods such as \citet{spectral_detuning} also allows restoration of the weights.

\begin{figure*}[t]
    \centering
    \includegraphics[width=1\linewidth]{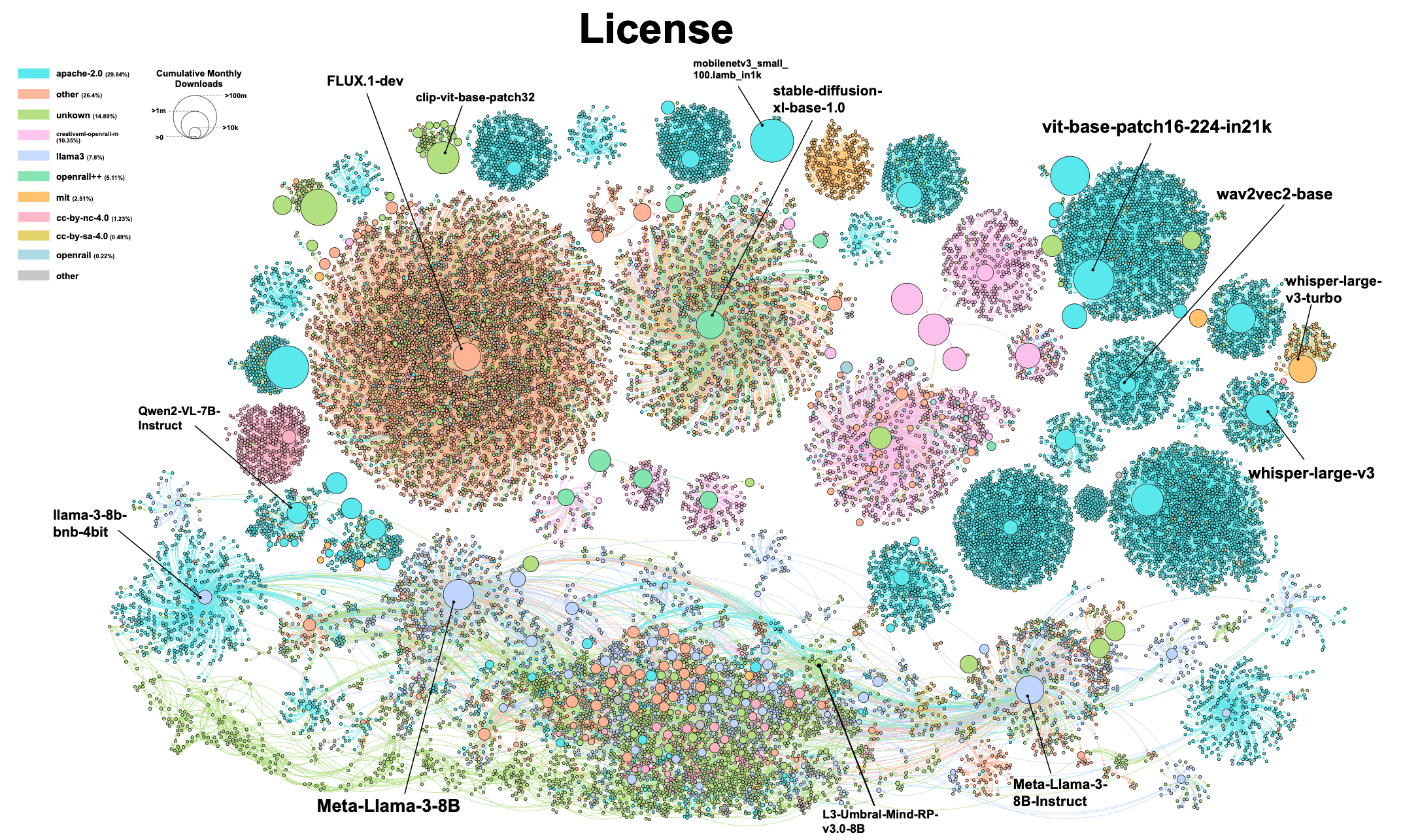}  \\ 
        \includegraphics[width=1\linewidth]{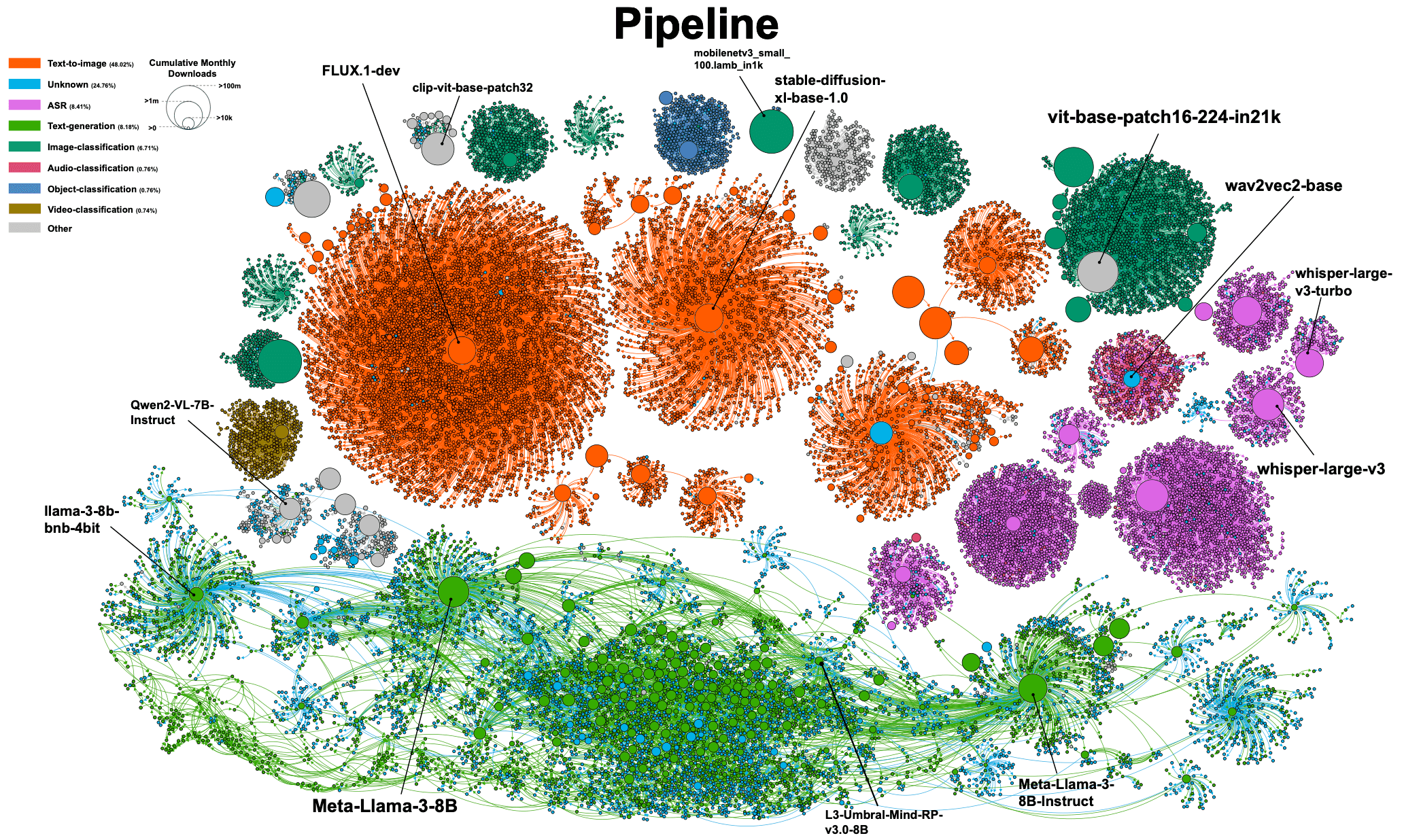}  \\ 
    \caption{\textit{\textbf{Visualizing other node features}}}
    \label{fig:other_att}
\end{figure*}

\begin{figure}[t]
\centering
\begin{subfigure}[b]{0.485\textwidth}
    \centering
       \includegraphics[width=0.9\linewidth]{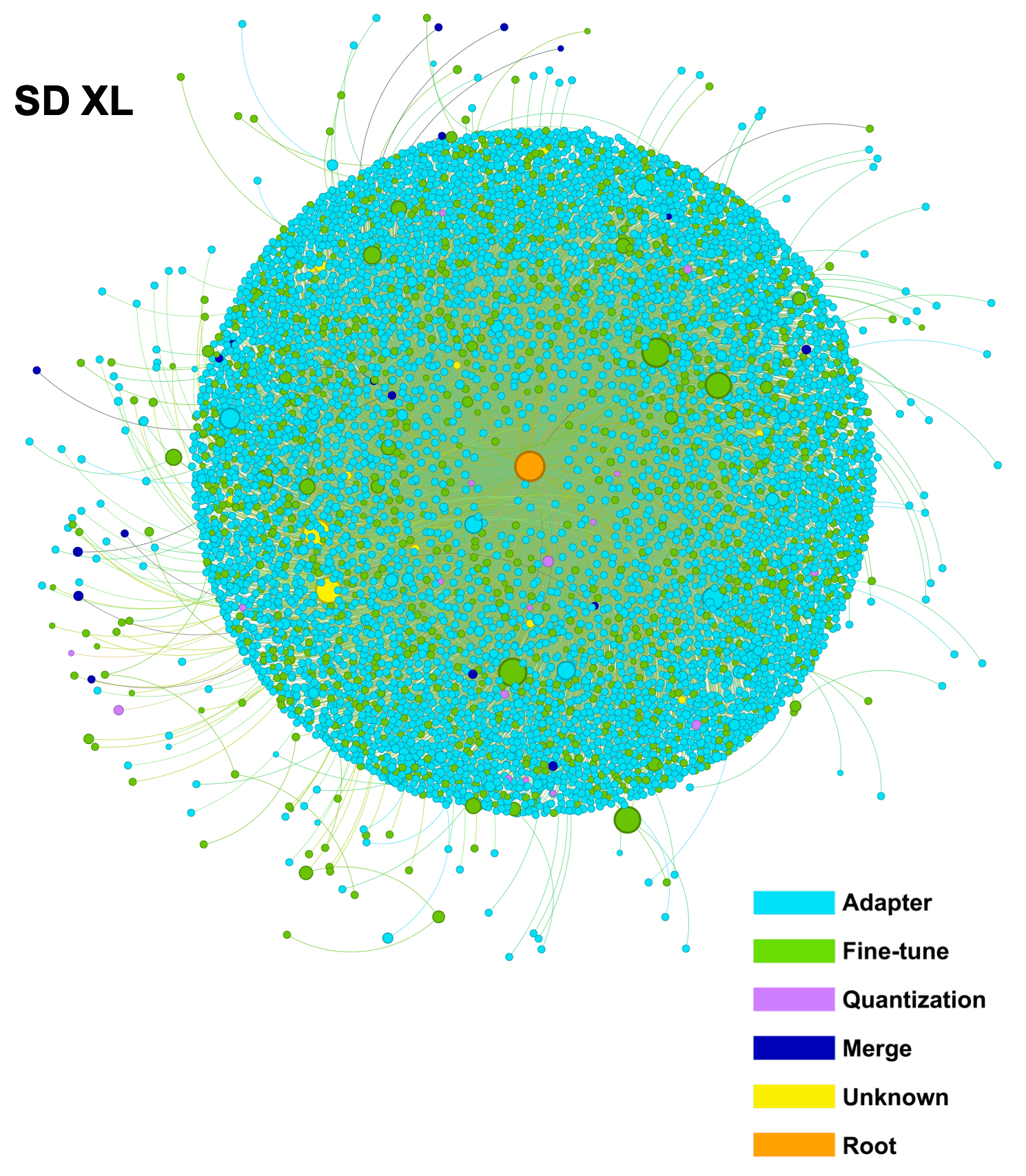}  \\
    \caption{\textit{\textbf{Individual connected components - 3/6}}}
    \label{fig:cc3}
\end{subfigure}
\hfill
\begin{subfigure}[b]{0.485\textwidth}
    \centering
    \includegraphics[width=0.9\linewidth]{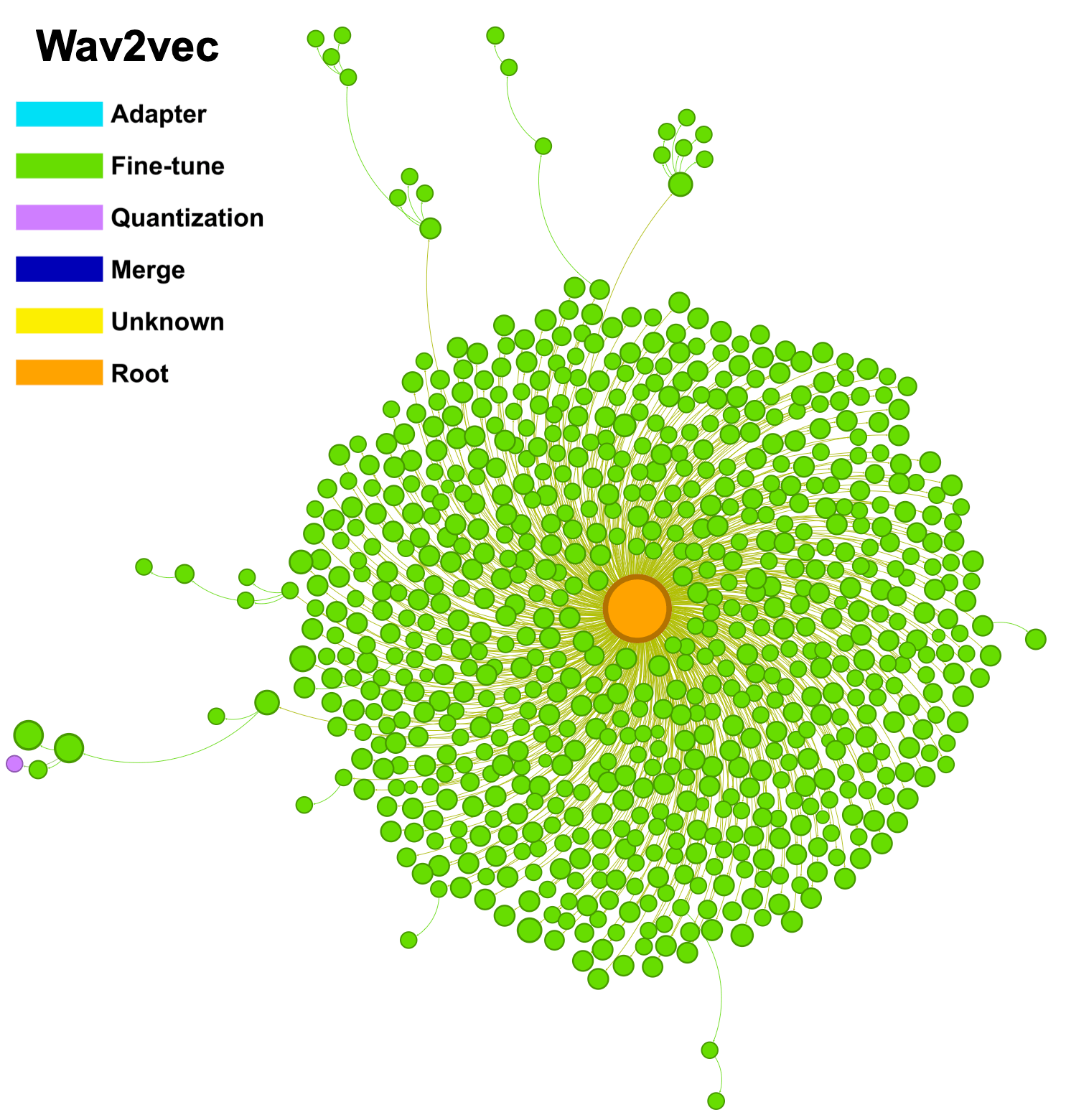}  \\  
    \caption{\textit{\textbf{Individual connected components - 4/6}}}
    \label{fig:cc4}
\end{subfigure}
\end{figure}

\begin{figure*}
    \centering
    \includegraphics[width=1\linewidth]{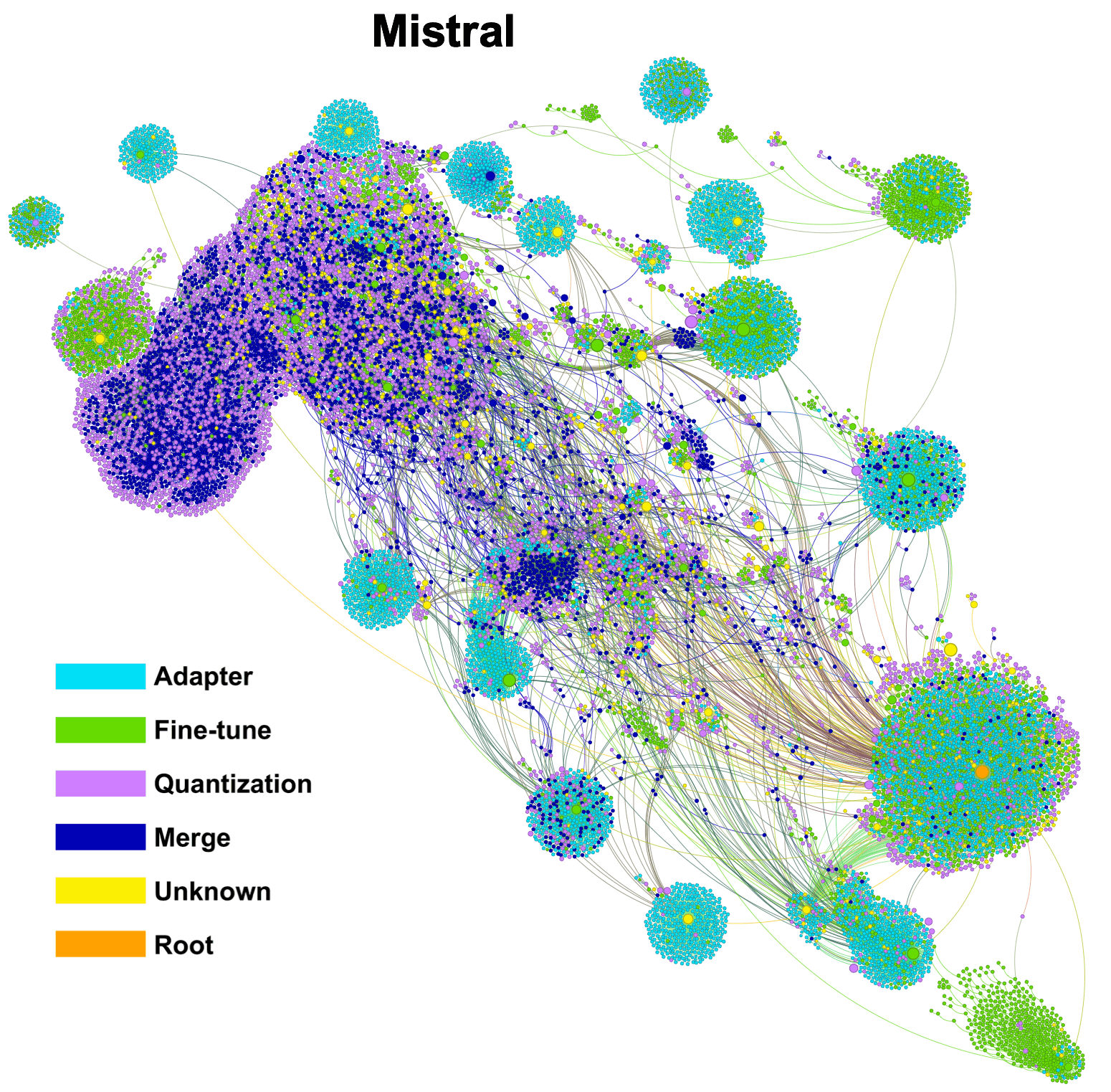}  \\ 
    \caption{\textit{\textbf{Individual connected components - 5/6}}}
    \label{fig:cc5}
\end{figure*}

\begin{figure}
    \centering  
    \includegraphics[width=\linewidth]{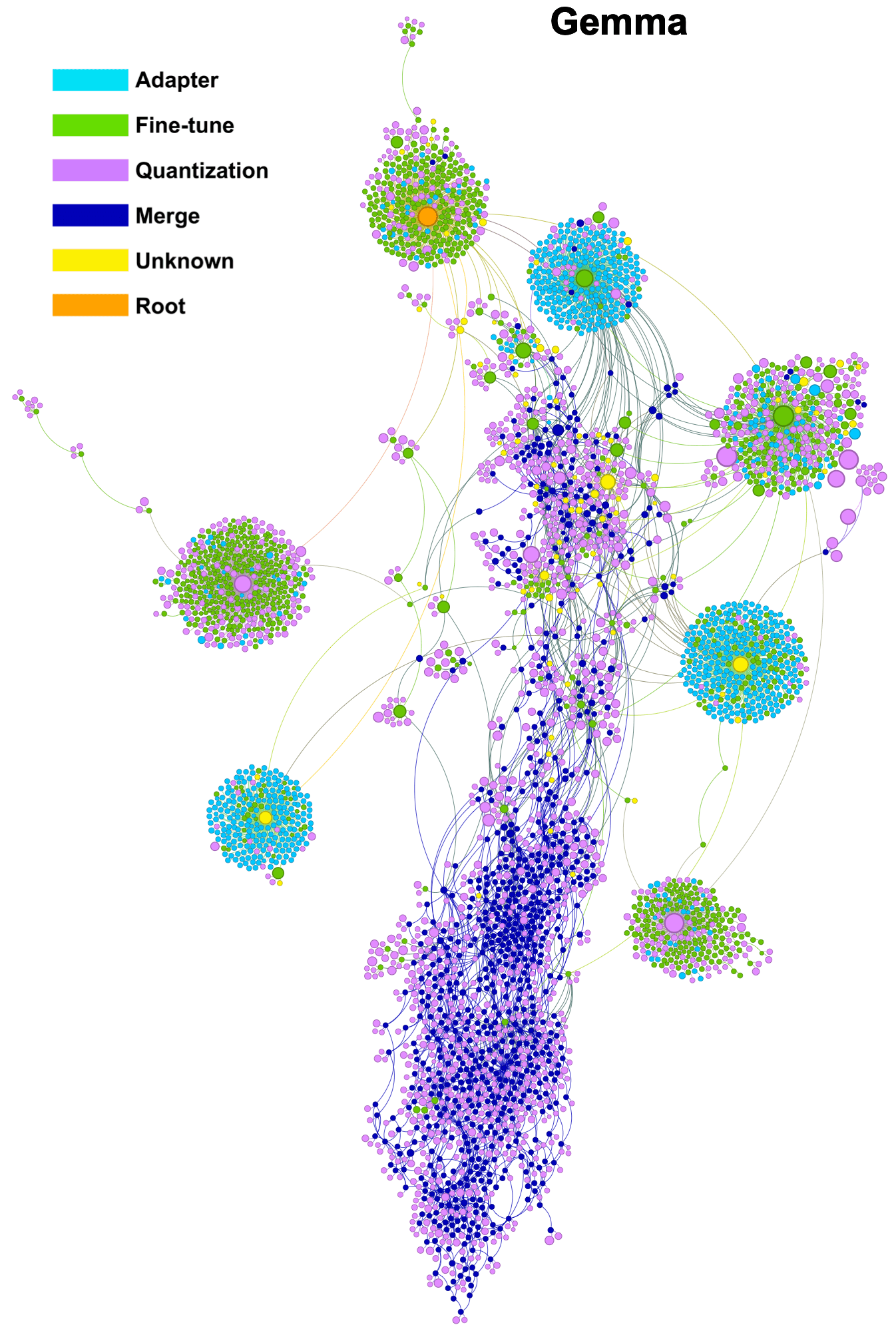}  \\      
    \caption{\textit{\textbf{Individual connected components - 6/6}}}
    \label{fig:cc6}
\end{figure}

\noindent \emph{Additional visualizations} In \cref{fig:other_att}, we visualize the same atlas regions shown in \cref{fig:motivation}, but colored according to different node features, specifically, license type and pipeline tag. In \cref{fig:cc1,fig:cc2,fig:cc3,fig:cc4,fig:cc5,fig:cc6}, we provide detailed visualizations of individual connected components. Figures have been compressed to reduce file size.

\section{Additional open challenges in Model Atlas charting}
\label{app:additional_open_challenges}
\noindent \emph{Meta-ML research.}  
This paper presents a simple example of meta-ML research through structural and visual analysis of the Model Atlas. However, extracting deeper insights will require more sophisticated methods. One direction involves visual and interactive exploration, comparing regions of the atlas by architecture family, task domain, training patterns, or performance metrics. Another involves training models on the atlas structure itself, for example, using graph neural networks (GNNs) to predict missing attributes. Beyond analysis, the Model Atlas can be extended by the broader community through the addition of new layers of node and edge features. Researchers in interpretability, for instance, might annotate models with salient neurons or circuits, while robustness researchers could contribute metrics such as adversarial susceptibility or failure modes. These enriched features would support downstream meta-ML tasks, enabling comparisons of interpretability across architectures or correlations between robustness and lineage.

\noindent \emph{Model charting datasets.} As our goal is to scale model charting to real-world scenarios, it is essential to use datasets drawn directly from real-world repositories. However, most existing dataset papers \citep{schurholt2022model,schurholt2025model,falk2025model} and methods that introduce datasets \citep{probex,mother,spectral_detuning,schurholt2022hyper,lol} rely on training their own model collections.  Curating datasets and benchmarks from models hosted on Hugging Face offers a convenient and scalable alternative. In fact, using the already-charted Model Atlas, one can assemble large datasets of in-the-wild models directly from Hugging Face. We include two curated datasets with this paper for graph-level model attribute prediction and charting tasks (see \cref{app:dataset_details} for details).

\section{Charting the atlas structure}
\label{sec:charting_method}
\begin{wrapfigure}[12]{R}{0.33\textwidth}
    \centering
    \includegraphics[width=\linewidth]{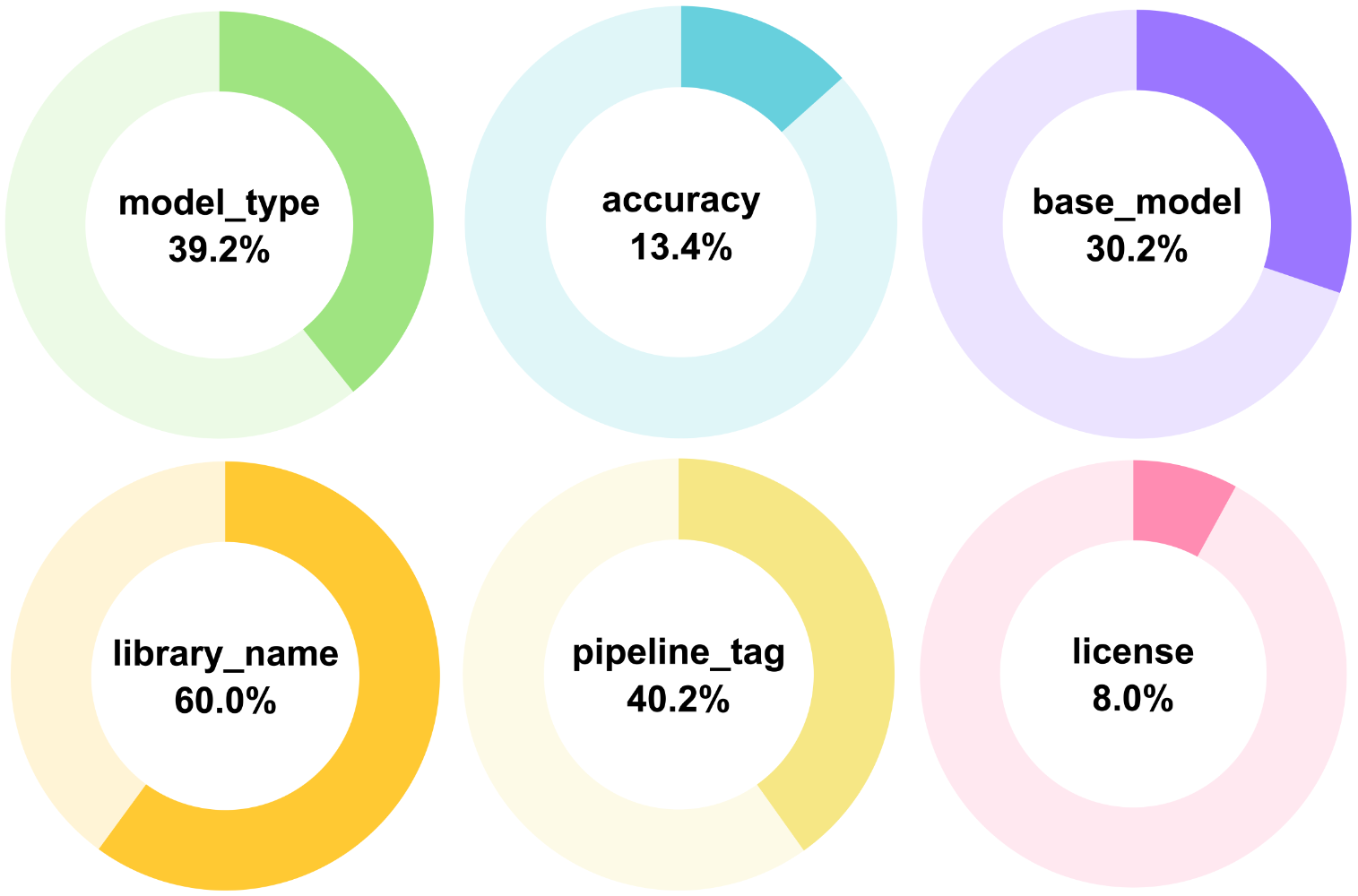}
    \caption{\textit{\textbf{Documentation level in Hugging Face}}}
    \label{fig:documentation_all}
\end{wrapfigure}

While the preceding section highlighted the importance of the Model Atlas, in practice, over 60\% of it is unknown. This is primarily because model uploaders frequently do not provide parent model information, as illustrated in \cref{fig:documentation_all}. Atlas charting aims to recover the missing edges. Recent methods \citep{mother,neural_phylogeny,phylolm,neural_lineage} tackled some aspects of this task, but were not applied to fully in-the-wild settings. As such, some of their key assumptions (e.g., tree-like structures) do not hold in real-world repos. Additionally, some methods \citep{phylolm} require running each model on multiple inputs, which is computationally infeasible for millions of models. We introduce an approach for in-the-wild charting. 

Charting typically begins by measuring pairwise model distances, which requires a representation for each model as well as a distance function. Common model representations include: weights \citep{mother,phylolm,neural_lineage,neural_phylogeny}, activations \citep{phylolm}, gradients \citep{neural_lineage}, outputs \citep{probelog,lu2023content}, metadata \citep{luo2024stylus} or a combination of the above \citep{neural_lineage}. We represent each model $i$ by its weights $w_i$. We measure the distance between two models using the Euclidean distance of their representations: $D_{ij} = \|w_i - w_j\|^2$.

\begin{wrapfigure}[18]{R}{0.5\textwidth}
\centering
   \includegraphics[width=1.0\linewidth]{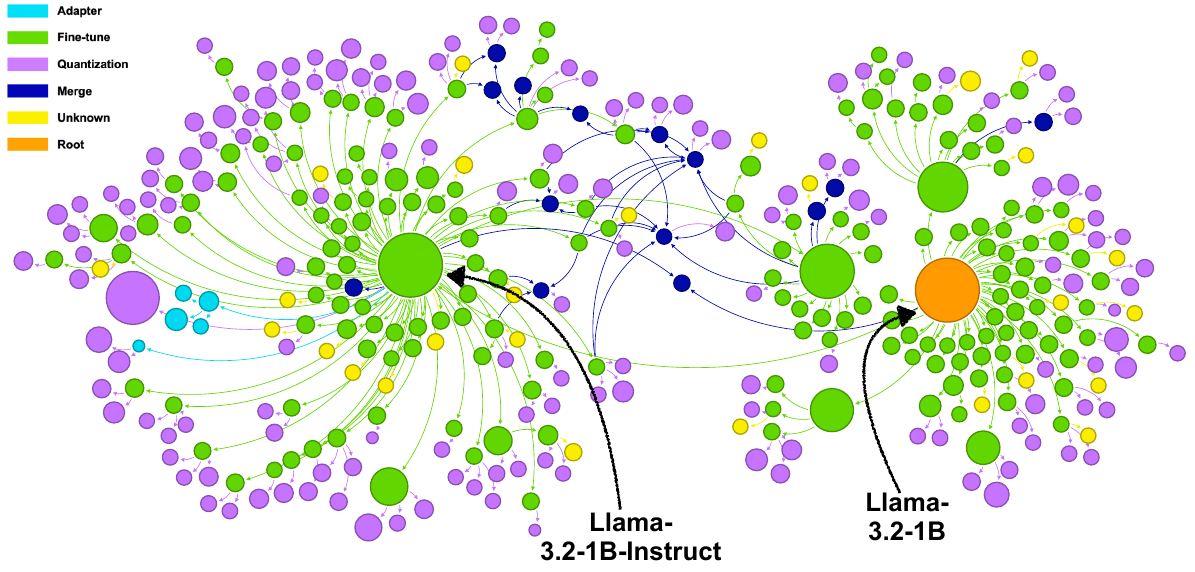}
            \caption{\textit{\textbf{Quantizations are leaves:}} 
             Our analysis of over 400,000 documented model relationships reveals that $99.41\%$ of quantized models are leaf nodes. This figure shows this for a subset of the Llama-based models. Indeed, quantized models (magenta) are nearly always leaf nodes, corroborating the statistical finding.}
            \label{fig:quantization_constraint}
\end{wrapfigure}

The core challenge is charting the atlas structure from this distance matrix. While previous approaches used simple, generic solutions, such as greedy nearest neighbor or minimal directed spanning tree, this does not generalize to in-the-wild Model Atlases. The reasons for this include: i) models with the nearest weight distance are not always connected by an edge, ii) the Model Atlas structure is generally not a tree. Instead, we propose a set of charting rules motivated by special patterns in model repos: duplication, quantization near-duplications, checkpoint trajectories, hyperparameter search, and merging. Identifying and handling these patterns can dramatically improve charting accuracy.

\noindent \textbf{Naive strategy.} Our preliminary strategy first splits models into non-overlapping subsets using a standard clustering algorithm on the weight distance matrix. Previous works \cite{mother,probex,gueta2023knowledge} showed that each subset corresponds to the nodes within a single connected component of the atlas DAG, i.e., we can recover the structure of each connected component separately. The naive strategy first assigns source nodes (this will be elaborated on below), then, at each step, it looks for the unassigned node with the shortest distance to the assigned nodes, and connects the two with a directed edge. The algorithm stops when there are no unassigned nodes.

\wrapassumpbox{6}{R}{0.46\textwidth}{Duplication Assumption}{Exact duplicates and quantized versions of models are treated as leaf nodes in the atlas.}

\noindent \textbf{Duplicates and near-duplicates.} Real-world model repositories contain many duplicates and near-duplicates. Exact duplicates occur when users download a popular base model and re-upload it without modification. In this, both the original and duplicates have \textit{identical} distances to all models. This makes the greedy distance-based algorithm arbitrarily decide between the true parent and its duplicates, which obviously reduces accuracy. To mitigate this, we identify models with zero distance as exact duplicates. We retain a single representative instance, designating duplicates as leaves.

Model quantization compresses models to reduce memory and storage as well as accelerate their inference. It transforms a model into a near-duplicate, as the values of each weight are typically very similar yet not identical to the original one. This creates a similar ambiguity to exact duplicate models, as the distance of the original and quantized model to any other model are almost the same. Moreover, many models have multiple quantizations, which greatly increases this ambiguity.

To overcome this ambiguity, we identify a pattern of quantization models from real-world atlases. Our analysis reveals that $99.41\%$ of quantized models on HF are leaf nodes, i.e., they have no child models. See \cref{fig:quantization_constraint} for an illustration on the Llama 3.2-1B CC. Intuitively, unquantized models have better performance than their quantized versions, and practitioners typically use the highest-performing models for further fine-tuning or merging. Fortunately, detecting quantization is often straightforward. Quantized models have reduced precision data types (e.g., int8, float16 instead of float32) and fewer unique weight values. Therefore, we simply identify quantized models and designate them as leaves.

\wrapassumpbox{9}{R}{0.5\textwidth}{Temporal Dynamics Assumption}{Models with earlier timestamps are assumed to be topologically higher in the atlas. For snake patterns, the parent is the closest preceding model; for fan patterns, it is the earliest model among the top $K$ weight-based neighbors.}

\begin{wrapfigure}[23]{R}{0.51\textwidth}
\centering
            \includegraphics[width=\linewidth]{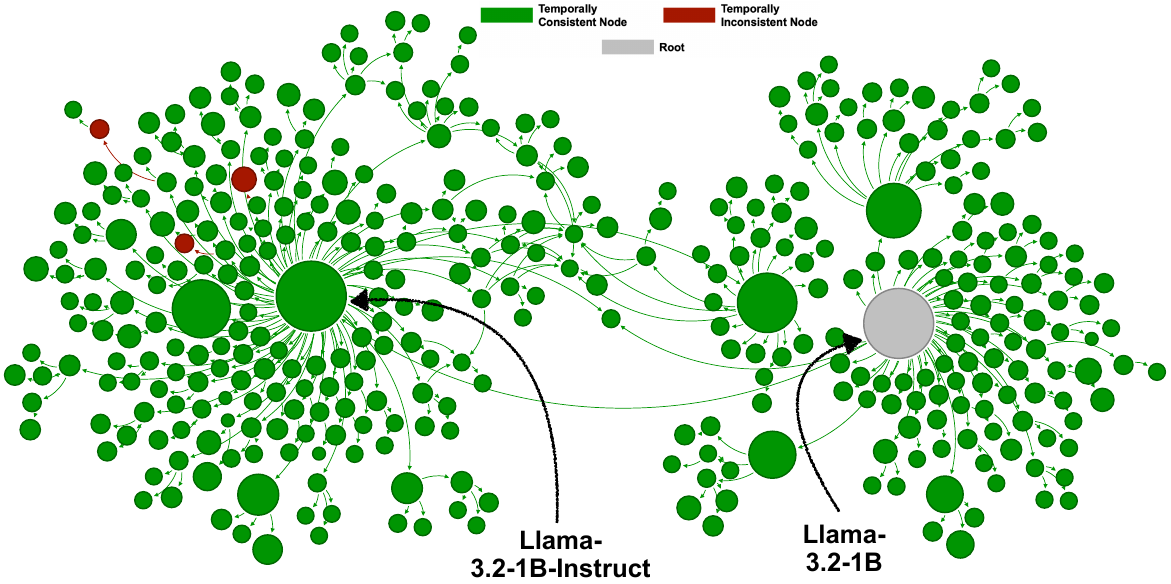}
            \caption{\textbf{\textit{Temporal dynamics indicate edge directionality:}} We analyzed over 400,000 documented model relationships and observed that in $99.73\%$ of cases, earlier upload times correlate with topologically higher positions in the DAG. Here, we visualize this trend on a subset of the Llama model family. Green nodes indicate models where earlier upload times align with topological order, while red nodes represent exceptions to this trend. The source (in gray)  vacuously satisfied this assumption. It is clear that nearly all nodes satisfy our assumption.}
            \label{fig:temporal_constraint}
\end{wrapfigure}

\noindent \textbf{Temporal dynamics for inferring fine structure.} While the weight distance between models is inversely correlated to the likelihood of having an edge between them, it does not reveal the direction of the edge. Previous approaches predicted edge direction using supervised learning methods or by unlearned metrics, such as gradients or weight kurtosis. However, we find that these methods often do not generalize effectively to real-world model repositories. 
A key advantage of real-world repositories, unused by prior works, is the availability of model creation or upload timestamps. We find that the vast majority of parent models have earlier timestamps than their children, specifically, in the HF model repository, this occurs for $99.73\%$ of all observed parent-child pairs. We visualize this phenomenon for the Llama-3.2-1B CC in \cref{fig:temporal_constraint}. This is a powerful temporal constraint on atlas structure, and in particular identifies the source nodes as the earliest ones.

\begin{wrapfigure}[21]{R}{0.51\textwidth}
\centering
            \includegraphics[width=\linewidth]{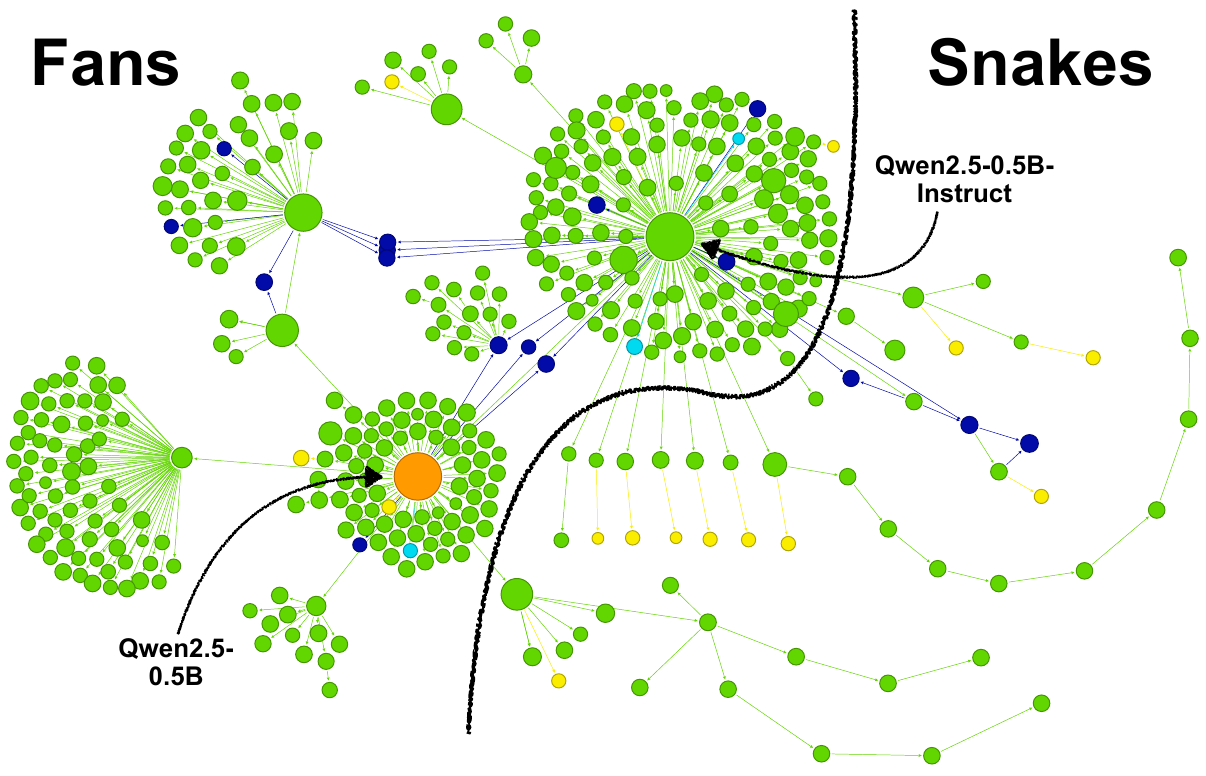}
            \caption{\textbf{\textit{Snake vs. Fan patterns:}} Snake patterns often arise from sequential training checkpoints, while fan patterns typically result from hyperparameter sweeps. In both structures, the model weight variance is low. However, in snake patterns, the weight distance has a high correlation with model upload time, whereas in fan patterns, the correlation is lower. Note colors are the same as \cref{fig:quantization_constraint}}
            \label{fig:snakes_vs_fans}
\end{wrapfigure}

Another limitation of the weight distance is that it may have limited sensitivity in fine-grained atlas structures. Two common scenarios that challenge charting methods based on weight distance are hyperparameter sweeps and checkpoint trajectories. In hyperparameter sweeps, multiple models are trained from a single parent, each with different hyperparameters. Minor hyperparameter changes can result in sibling models having nearer weight distances to each other than to their common parent, defying the nearest neighbor algorithm. Conversely, checkpoint trajectories represent a sequence of models saved during a single training run. In this case, edges should connect consecutive checkpoints in a chain-like manner, rather than connecting each checkpoint directly to the initial model (as in the hyperparameter sweep scenario). We term the former pattern a \q{fan} pattern and the latter a \q{snake} pattern, illustrated in \cref{fig:snakes_vs_fans}, and find that the weight distance often confuses these two patterns.

Our key observation is that the temporal model weight evolution can discriminate between these patterns. In snake patterns, temporal proximity strongly correlates with weight proximity, due to the sequential evolution. However, this is not the case in fan patterns, where the closest siblings are not necessarily the closest in time. This leads to a simple yet effective decision rule. If the weight distance of a model to its K nearest neighbors is highly correlated to their absolute temporal distance, we classify the pattern as a snake; otherwise, a fan. Then, we connect the node to one of its nearest neighbors depending on the detected structure. In snakes, we connect the node to its first nearest neighbor based on weight distance, while in fans, we connect the node to its oldest nearest neighbor, aiming for the fan's origin.

\wrapassumpbox{6}{R}{0.35\textwidth}{Structural Assumption}{The atlas structure is a non-tree directed acyclic graph (DAG).}

\noindent \textbf{Merged models.} Prior work assumes that models lie in a directed tree, however, as merged models have multiple parents, they have more than one incoming edge. This cannot be described by trees, but can be described by non-tree DAGs. We found that many merged models are created using a few popular libraries, which typically document the parent nodes in the created model card, hence, the multiple incoming edges are often known. The challenge left is to chart a DAG instead of a tree. We propose a greedy charting algorithm. We first sort all nodes by upload times. Then, we process each node in temporal order, predict its parents, and add those edges to the graph. In the case where the parents are known, we use these edges instead of predicting them. The algorithm stops where there are no more nodes to process. The runtime complexity is $O(n^2)$, where $n$ is the number of models. It is much faster than \citep{mother} (which is typically $O(n^3)$). In practice, we recover graphs with thousands of nodes in seconds.

The most expensive part of the algorithm in terms of compute and storage is the distance matrix calculation, as it scales with the number of network weights. Therefore, we propose to represent each model by subsampling the full weight representation to a much smaller number of neurons. Specifically, we find that keeping just $100$ is typically sufficient for significantly speeding up the runtime while keeping accuracy nearly unchanged.

\newpage \noindent \textbf{Full algorithm.} We present the full algorithm in python-like pseudo code in \cref{alg:addition}. The input is a set of models $M$, each with weights $w$, creation time $t$, known parents $P$ (or $P=None$), and whether it is quantized $q$. 
The algorithm has $3$ hyperparameters: $K$ - the number of neighbors, $K_{th},\rho_{th}$ - thresholds for snake-fan classification. The output of the algorithm is the predicted parents for each node $m.P$. Note that we can run the algorithm online with a small modification, by computing the distance function and kNN as new models arrive. See \cref{app:implementation_details} for implementation.

\begin{wrapfigure}[25]{R}{0.5\textwidth}
\captionsetup{type=algorithm}
\rule{\linewidth}{1pt} 
   \captionof{algorithm}{Atlas structure recovery (psuedo-code)}
   \label{alg:addition}
\hrule

    \definecolor{codeblue}{rgb}{0.25,0.5,0.5}
    \definecolor{codekw}{rgb}{0.85, 0.18, 0.50}
    \lstset{
  backgroundcolor=\color{white},
  basicstyle=\fontsize{6.5pt}{6.5pt}\ttfamily\selectfont,
  columns=fullflexible,
  breaklines=true,
  captionpos=b,
  commentstyle=\fontsize{6.5pt}{6.5pt}\color{codeblue},
  keywordstyle=\fontsize{6.5pt}{6.5pt}\color{codekw},
}
\begin{lstlisting}[language=python]
# Input: M = {(w, q, t, P)}, K, K_th, rho_th
# w = weights, t = creation time, P = known parents
M = M.sort(key=m.t) # Sort M by timestamp t
D = compute_distance_matrix(M)
D.diag = inf

for i, m in enumerate(M):
    if is_quantized(m.w):  # Set quantized model as leaf node
        D[i, :] = inf
    if m.P:  # Use existing parents if available
        continue
    for j in range(i+1, len(M)): # Deduplication
        if D[i, j] == 0:
            D[j, :] = inf
            M[j].P = M[i].P
    
    # Find k-NN
    k_ind = argsort(D[:, i])[1:K]
    temporal_k_ind = argsort(D[i+1:, i])[1:K]
    k_dist = D[k_ind, i]
    k_times = [M[k].t for k in k_ind]
    temporal_k_times = [M[k].t for k in temporal_k_ind]
    
    
    # Check spread of distances
    if k_dist[-1] - k_dist[0] > K_th:
        m.P = temporal_k_ind[0]
    else:
        corr = correlation(k_dist, abs(k_times - m.t))
        if corr > rho_th: # Snake
            m.P = temporal_k_ind[0]
        else: # Fan
            m.P = temporal_k_ind[argmin(temporal_k_times)]
\end{lstlisting}
\hrule
\end{wrapfigure}

\subsection{Charting results}
\label{sec:experiments}

\noindent \textbf{Datasets.} We created an atlas charting dataset based on the ground truth model relation found on the \q{hub-stats}\footnote{\url{huggingface.co/datasets/cfahlgren1/hub-stats}} dataset. It consists of 3 atlas connected components: \texttt{Qwen2.5-0.5B} (Qwen), \texttt{Llama-3.2-1B} (Llama), and \texttt{Stable-Diffusion-2} (SD). We also created another attribute prediction dataset. Its task is to predict 6 attributes: \texttt{accuracy}, \texttt{pipeline\_tag}, \texttt{library\_name}, \texttt{model\_type}, \texttt{license}, \texttt{relation\_type}. The ground truth was obtained by a combination of model metadata and running an LLM on the model cards. We detail the dataset creation process in \cref{app:dataset_details} and will publicly release the datasets.

\noindent \textbf{Baselines.} We compared a weight-agnostic classical method and a newer method that depends on weights. \textit{MoTHer \citep{mother}} - a recent weight-dependent method. Importantly, it predicts edge direction using weight kurtosis and relies on the structure being a tree. We could not scale other weight-based methods to operate on our in-the-wild dataset. We also included the following structure-only baselines: \textit{random} edge assignment, \textit{majority vote} (assigning all nodes to the parent with the highest out-degree), and \textit{Price’s model \citep{prices_model}}. The latter is a preferential attachment algorithm \citep{BA_model}, popular for citation networks, that assigns edge probabilities based on node out-degree. It is essentially a probabilistic version of majority vote that allows for new \q{hub} formation. Note that all baselines (except random) assume the structure is a tree, not a DAG. To accommodate the majority and Price's models, we assumed a graph stem with known edges, containing $10\%$ of the nodes in the connected component. We evaluated all methods by their accuracy on the remaining $90\%$ of models.

\noindent \textbf{Baseline comparison.} We use our method to chart three atlas connected components. As we can see in \cref{tab:recovery_results}, the baselines perform better than random but still poorly. The majority approach models the graph as depth-1 centered around the root. This obtains low but non-trivial results, as this model is too simplistic. Price's model is more powerful and can create a more realistic looking structure. However, it fails as it is data agnostic and connects the wrong nodes.  MoTHer performs better on some DAGs, but its directional prediction is weaker and it lacks our other priors, resulting in lower performance. 

\begin{table}[t]
    \centering
    \begin{minipage}[t]{0.48\textwidth}
        \centering
        \caption{\textbf{\textit{Atlas recovery results:}} Our method outperforms the baselines by a significant margin, even for in-the-wild models.}
        \label{tab:recovery_results}
        \resizebox{\linewidth}{!}{%
        \begin{tabular}{lccc}
            \textbf{Method} & \textbf{Qwen} & \textbf{Llama} & \textbf{SD} \\
            \toprule
            Random - root & 1.77 & 1.84 & 0.22 \\
            Random & 0.98 & 0.67 & 0.75 \\
            Majority & 15.03 & 25.00 & 36.75 \\
            Price's \citep{prices_model} & 2.28 & 5.08 & 8.50 \\
            MoTHer \citep{mother} & 32.81 & 19.32 & 50.51 \\
            Ours & \textbf{78.87} & \textbf{80.44} & \textbf{85.10} \\
        \end{tabular}
        }
    \end{minipage}\hfill
    \begin{minipage}[t]{0.48\textwidth}
        \centering
        \caption{\textbf{\textit{Subtractive ablation:}} We remove each of our assumptions individually, indeed, we see that each one contributes to our final high recovery accuracy.}
        \label{tab:recovery_ablation}
        \resizebox{\linewidth}{!}{%
        \begin{tabular}{llccc}
            \multicolumn{2}{c}{\textbf{Method}} & \textbf{Qwen} & \textbf{Llama} & \textbf{SD} \\
            \toprule
            \multirow{5}{*}{\rotatebox[origin=c]{90}{Ours}} 
            & $-$ Greedy Alg. & 77.34 & 78.98 & Failed \\
            & $-$ Quantization & 70.57 & 36.59 & 84.85\\
            & $-$ Deduplication & 67.85 & 75.28 & \textbf{88.76} \\
            & $-$ Temporal Consistency & 75.47 & 80.13 & 80.86 \\
            & $-$ Fans vs. Snakes & 75.09 & 76.03 & 71.72 \\
            & $-$ Merges & 76.95 & 76.61 & 84.85 \\
            \midrule
            \multicolumn{2}{c}{\textbf{Ours}} & \textbf{78.87} & \textbf{80.44} & 85.10 \\
        \end{tabular}
        }
    \end{minipage}
\end{table}

\section{Implementation details}
\label{app:implementation_details}
To visualize the atlas, we use the open-source software Gephi \citep{bastian2009gephi}. We set $K=5$ nearest neighbors in all DAGs and determine the snake correlation threshold ($\rho_{\text{th}}$) dynamically for each DAG as the $60$th percentile of all node correlations. The distance spreading correlation threshold ($K_{\text{th}}$) is fixed at $0.05$. 

Additionally, in all our experiments, we use $100$ random neurons as model features. For Stable Diffusion, we select these neurons exclusively from the attention layers, as these layers typically undergo LoRA fine-tuning. In all DAGs, we allocate $10\%$ of the actual DAG for training and reserve the remaining $90\%$ as a test set. These hyperparameters are applied consistently across all three tested DAG structure recovery settings.

\section{Dataset details}
\label{app:dataset_details}
We construct two datasets from the Hugging Face \q{hub-stats} dataset\footnote{\url{https://huggingface.co/datasets/cfahlgren1/hub-stats}}: (i) Metadata Imputation and (ii) DAG Recovery. We will make our dataset publicly available on Hugging Face. To process the data, we first used the \texttt{base\_model} attribute to group models into connected components. Since many models lacked this attribute, we manually inspected the largest 800 connected components and filled in the missing values. This step helped merge large components with incomplete information and reduced data noise.

The metadata imputation dataset includes all 1.3 million models. The tasks involve imputing the following attributes: \texttt{pipeline\_tag}, \texttt{library\_name}, \texttt{model\_type}, \texttt{license}, and \texttt{relation\_type}, as well as estimating model evaluation metrics. To obtain the evaluation metrics, we downloaded model cards and extracted the metrics using \textit{deepseek-ai/DeepSeek-R1-Distill-Qwen-7B} in a zero-shot setting on models from \textit{mistralai/Mistral-7B-v0.1} CC.

To construct the charting dataset, we created a graph of all models on Hugging Face. Using the \texttt{base\_model} attribute, we established ground truth edges between models that reference each other. We removed all nodes with no parent or child (i.e., CCs with a single model), reducing the initial 1.3 million models to approximately 400,000 models across different connected components. We defined sources as nodes with no incoming edges and considered weakly connected components, ensuring that each CC has a single source node.

To facilitate model downloads, we pruned the CCs by randomly sampling $K$ leaf nodes if a node had more than $K$ leaf nodes, filtering out the rest. This allowed us to download hundreds rather than tens of thousands of models. We set $K=3$. If a download failed, we removed the model from the ground truth atlas.

During the download process, quantized model weights required special handling to convert them back into workable weight files. The conversion was based on the tensor type, which indicates whether a model has undergone quantization. While reviewing the Hugging Face model cards, we discovered multiple tagging inconsistencies; some models were incorrectly labeled as quantized, while others were missing the label. We propose using the tensor type as a reliable indicator of quantization status.

\end{document}